\documentclass[preprint,authoryear,3p, 11pt]{elsarticle}
\usepackage{amssymb}
\usepackage[utf8]{inputenc} % allow utf-8 input
\usepackage[T1]{fontenc}    % use 8-bit T1 fonts
\usepackage{hyperref}       % hyperlinks
\usepackage{url}            % simple URL typesetting
\usepackage{booktabs}       % professional-quality tables
\usepackage{amsfonts}       % blackboard math symbols
\usepackage{nicefrac}       % compact symbols for 1/2, etc.
\usepackage{microtype}      % microtypography
\usepackage{lipsum}

\usepackage{enumitem}
\usepackage{bbm}
\usepackage{physics}
\usepackage{bm}
\usepackage[short]{optidef}
\usepackage{booktabs}       % professional-quality tables

\usepackage{amsmath}
\usepackage{amssymb}
\usepackage{xfrac}
\usepackage{acronym}
\usepackage{subfig,graphicx}
\usepackage{algorithmic}
\usepackage[ruled,vlined]{algorithm2e}
\SetKwInput{KwInput}{Input}                % Set the Input
\SetKwInput{KwOutput}{Output} 
\usepackage{multirow}
\usepackage{mathtools}

\usepackage{caption}
\usepackage{subcaption}

\usepackage{tikz}
\usetikzlibrary{calc,patterns,arrows,shapes.arrows,intersections}

\hypersetup{pdfauthor={Name}}
\usepackage[space]{grffile} % For spaces in paths
\usepackage{etoolbox} % For spaces in paths
\makeatletter % For spaces in paths
\patchcmd\Gread@eps{\@inputcheck#1 }{\@inputcheck"#1"\relax}{}{}
\makeatother
\journal{European Journal of Operational Research}
\begin{document}

\begin{frontmatter}

%% Title, authors and addresses

\title{A Machine Learning Approach to Two-Stage Adaptive Robust Optimization}

%% use optional labels to link authors explicitly to addresses:
\author[1]{Dimitris Bertsimas\corref{cor1}}
\ead{dbertsim@mit.edu}

\author[2]{Cheol Woo Kim }
\ead{acwkim@mit.edu}

\cortext[cor1]{Corresponding author}

 % \affiliation[1]{organization={Sloan School of Management, Massachusetts Institute of Technology}, 
 %                 addressline={100 Main Street},
 %                 city={Cambridge}, 
 %                postcode={02142}, 
 %                 country={United States}}

\address[1]{Sloan School of Management, Massachusetts Institute of Technology, 100 Main Street, Cambridge, 02142, United States}

 % \affiliation[2]{organization={Operations Research Center, Massachusetts Institute of Technology},
 %                 addressline={1 Amherst Street}, 
 %                 city={Cambridge},
 %                 postcode={02142}, 
 %                 country={United States}}
\address[2]{Operations Research Center, Massachusetts Institute of Technology, 1 Amherst Street, Cambridge, 02142, United States}

\begin{abstract}
We propose an approach based on machine learning to solve two-stage  linear adaptive  robust optimization (ARO) problems with binary here-and-now variables and polyhedral uncertainty sets. We encode the optimal here-and-now decisions, the worst-case scenarios associated with the optimal here-and-now decisions, and the optimal wait-and-see decisions into what we denote as the strategy. We solve multiple similar ARO instances in advance using the column and constraint generation algorithm and extract the optimal strategies to generate a training set. We train machine learning models that predict high-quality strategies for the here-and-now decisions, the worst-case scenarios associated with the optimal here-and-now decisions, and the wait-and-see decisions. {\color{black} The models can be applied to problems with varying dimensions. We also introduce novel methods to expedite training data generation and reduce the number of different target classes the machine learning algorithm needs to be trained on.} We apply the proposed approach to the facility location, the multi-item inventory control and the unit commitment problems. Our approach solves ARO problems drastically faster than the state-of-the-art algorithms with high accuracy. 

\end{abstract}

%%Graphical abstract

%%Research highlights

\begin{keyword}
Machine Learning \sep Adaptive Robust Optimization  \sep Articifial Intelligence
%% keywords here, in the form: keyword \sep keyword

%% PACS codes here, in the form: \PACS code \sep code

%% MSC codes here, in the form: \MSC code \sep code
%% or \MSC[2008] code \sep code (2000 is the default)

\end{keyword}

\tnotetext[1]{This manuscript has been accepted for publication in European Journal of Operational Research. The final version is available at: \url{https://doi.org/10.1016/j.ejor.2024.06.012}.}

\end{frontmatter}

%% \linenumbers

%% main text

%%%%%%%%%%%%%%%%%
%% SEC. INTRODUCTION
%%%%%%%%%%%%%%%%%
\section{Introduction}
\label{sec:intro}

Robust optimization (RO) has become increasingly popular as a method to account for parameter uncertainty. Compared to more conventional methods such as stochastic optimization, which can be computationally intensive in high dimensions, RO offers a significant computational advantage \citep{robusttext,Bertsimasrobusttext}.
%In contrast, robust optimization is tractable even in very large dimensions \citep{priceofrobust}, \citep{robustconvex},\citep{robusttext}.

Adaptive robust optimization (ARO) is an important extension of RO that allows certain decision variables, referred to as the wait-and-see variables, to be determined after the uncertainty is revealed. In ARO, the wait-and-see decisions are mathematically modeled as functions of uncertain parameters, enabling them to adapt to the realization of those parameters. ARO is particularly useful in multi-stage decision-making problems, where decision-makers may be uncertain about future parameter values, and where decisions may need to be made sequentially over time. Compared to RO, ARO provides greater modeling flexibility and often results in superior solutions that are better able to adapt to changing conditions \citep{adaptivefirst}. Application areas include energy \citep{Sun2014Wind, BertsimasSun}, inventory management \citep{inventory1, inventory2}, portfolio management \citep{portfolio}, machine scheduling \citep{COHEN202383} among many others. 
 
 Despite its many benefits, ARO poses significant computational challenges that distinguish it from RO. One of the primary challenges arises from the fact that ARO consists of infinite-dimensional optimization problems, as the wait-and-see variables are functional variables. To overcome this issue, approximation methods have been proposed that restrict the wait-and-see variables to a limited set of functions, such as affine functions \citep[Section~7]{Bertsimasrobusttext}. However, while these methods may be able to reformulate ARO into RO, there is no guarantee that the resulting approximation will be near-optimal or even feasible \citep[Section~5]{survey}. Other methods have been developed that can ensure near-optimal or even optimal solutions for ARO, including Benders Decomposition \citep{BertsimasSun}, Column and Constraint Generation (CCG) \citep{CCG}, scenario reduction \citep{GOERIGK2023529, Kai2023}, branch-and-bound \citep{Lefebvre2024} and Fourier-Motzkin Elimination \citep{adaptive}. These methods, however, may not scale well in high dimensions or assume specific structure on the uncertainties. Given the substantial computational burden of ARO, the application of ARO may be limited particularly in real-time settings where time and computational resources are severely constrained. In these domains where decisions need to be made within seconds or even milliseconds, opting for ARO can be often unviable. 
 
 To address this challenge, we propose a novel machine learning approach that can significantly reduce the computational burden associated with ARO. To generate a training set, we solve multiple ARO instances in advance using the CCG algorithm. Then, we train machine learning models to predict high-quality strategies for ARO problems that can simplify their solution process (exact definition of strategies will be presented in Section \ref{subsec:strategy}). To the best of our knowledge, our work is the first to harness the power of machine learning to tackle ARO. While our approach might involve heavy computational burden in the training phase, this investment enables us to attain remarkable speed-ups once the training is completed, outperforming state-of-the-art algorithms by several orders of magnitude. In practical terms, this means that ARO can now be solved in a matter of milliseconds. 
 
While previous work by \citet{Voice, Prune, OPT} explored machine learning techniques for solving mixed-integer convex optimization (MICO) problems, our work takes a leap by extending these methods to ARO. Their approach is to train a machine learning model that predicts the optimal strategy of a MICO problem, where the optimal strategy of a MICO instance is defined as the set of tight constraints and the set of binary variables that are equal to one.

{\color{black}
However, adapting these methods to the realm of ARO is not straightforward. ARO presents distinct mathematical structures and computational challenges compared to MICO. First, while MICO deals with finite-dimensional problems, ARO deals with infinite-dimensional ones. Consequently, it requires a different definition of the optimal strategies to encode the optimal solutions. Second, ARO, involving dynamic optimization problems, requires a comprehensive solution including not only here-and-now decisions but also worst-case scenarios associated with these decisions and subsequent wait-and-see decisions. Third, the computational complexity of solving ARO is substantially higher than MICO, posing challenges in training data generation. Finally, in the realm of ARO, the number of distinct target classes can grow significantly depending on the sampling strategy and the size of the uncertainty sets, making the prediction task difficult. Our proposed techniques effectively address these challenges inherent in ARO. Furthermore, unlike previous approaches, the machine learning models we train can handle problems with varying dimensions, thereby enhancing their versatility.

Our contributions are summarized as follows.}
\begin{enumerate}
    \item {\color{black} We propose a machine learning approach to solve two-stage linear ARO with binary here-and-now variables and polyhedral uncertainty sets. Our approach provides a comprehensive solution including high-quality here-and-now decisions, worst-case scenarios associated with these decisions and subsequent wait-and-see decisions. Moreover, the machine learning models can be applied to problems with varying dimensions.
    \item We propose a method to expedite the training data generation process, enhancing our approach's adaptability to shifting environments. 
    \item We propose a method to reduce the number of distinct target classes the machine learning model is trained on. This technique enables our approach to effectively address high-dimensional problems and accommodate large uncertainty sets. }
    \item We conduct a series of computational experiments involving both synthetic and real-world problems. The examples we test on include the facility location, the multi-item inventory control and the unit commitment problems. We demonstrate that we can obtain high-quality solutions using the proposed method. Notably, despite potentially lengthy training periods, the real-time application of our methodology dramatically outspeeds the state-of-the-art algorithms. In our experiments, we demonstrate a speed-up of more than 10 million times.
\end{enumerate}

The structure of this paper is as follows. In Section \ref{sec:2}, we briefly introduce ARO and explain how we solve a two-stage linear ARO problem with polyhedral uncertainty sets. We also demonstrate that this method may not scale to problems with high dimension. In Section \ref{sec:Machine Learning for Adaptive Optimizaion}, we develop a machine learning approach to solve two-stage ARO with binary here-and-now variables and polyhedral uncertainty sets. {\color{black} In Section \ref{sec:speedup}, we present a technique to accelerate training data generation.} In Section \ref{sec5}, we present an algorithm to reduce the number of different target classes. In Section \ref{sec:numerical experiment}, we present the results of numerical experiments. 

\paragraph{Notational Conventions} Throughout this paper, we use boldface letters to denote vectors and matrices. The $i_{th} $ entry of a vector $\bm{x}$ is denoted $x_i$ or $[\bm{x}]_i$, unless otherwise noted. For a positive integer $N$, we use $[N]$ to denote the set $\{i \in \mathbb{Z}: 1\leq i \leq N\}$. We use  $\bm{x}(\cdot)$ to denote a vector whose entries are real-valued functions.

\section{Two-stage Linear Adaptive Robust Optimization}
\label{sec:2}
In this section, we review two-stage linear ARO. We describe how to obtain the optimal here-and-now decisions, the worst-case scenarios associated with the optimal here-and-now decisions, and the optimal wait-and-see decisions using the CCG algorithm. We demonstrate numerically that this algorithm may not scale well with dimension. {\color{black} Additional computational analysis of the algorithm, including the impact of tolerance parameters and the choice of different initial points can be found in the Supplementary Material.}

\subsection{Problem Formulation}
Consider the two-stage ARO
\begin{alignat}{2}
&\underset{\bm{x},\bm{y}(\cdot)}\min\quad && \Bigg(\max_{\bm{d}\in\mathcal{D}}\bm{c}(\bm{d})^\top\bm{x} +\bm{b}^\top\bm{y}(\bm{d})\Bigg) \label{1} \\ 
& \text{s.t.}\quad &&  \bm{A}(\bm{d})\bm{x} + \bm{B}\bm{y} (\bm{d})\leq \bm{g},\quad \forall {\bm{d}\in\mathcal{D}}, \nonumber
\end{alignat}

 \noindent where $\bm{d}$ is a vector of uncertain parameters and $\mathcal{D}$ is a polyhedral uncertainty set. We also use the term scenario to refer to a specific realization of the uncertain parameter. $\bm{x}$ is the vector of here-and-now variables that represents the decisions that have to be made before the uncertainty is revealed.  $\bm{y}(\cdot)$ is the vector of wait-and-see variables, which is a function of $\bm{d}$. $\bm{A}(\bm{d})$ and $\bm{c}(\bm{d})$ are affine in $\bm{d}$. We assume $\bm{B}$ and $\bm{b}$ do not involve uncertainty, a condition commonly known as the fixed-recourse assumption. Without this assumption, solving an ARO instance becomes considerably more challenging \citep[Chapter~6\&7]{Bertsimasrobusttext}.
 
  The wait-and-see variables represent the decisions that can be made after the uncertainty is revealed.  This flexibility leads to less conservative and more realistic solutions compared to RO.  ARO results in better objective values because the wait-and-see variables can be decided based on actual realizations of the uncertain parameters, whereas in RO, conservative decisions must be made in advance. Moreover, ARO tolerates larger uncertainty levels than RO. In some cases, a RO problem can be infeasible if the uncertainty set is too large. However, by switching some of the decision variables to wait-and-see variables, the resulting ARO problem may become feasible \citep[Section~1\&5]{ adaptivefirst}.

As the wait-and-decision variable $\bm{y}(\cdot)$ is an arbitrary function, it represents an infinite-dimensional variable. To manage this challenge, a common approach is to constrain $\bm{y}(\cdot)$ to a family of parametric functions. Among the popular choices is the affine decision rule, where we assume that $\bm{y}(\bm{d}) = \bm{Pd} + \bm{z}$ for some $\bm{P}$ and $\bm{z}$. By substituting this expression back into problem \eqref{1}, $\bm{P}$ and $\bm{z}$ become finite vectors of decision variables alongside $\bm{x}$.

Another approximation method is to consider only a limited number of key scenarios from $\mathcal{D}$. In this approach, $\mathcal{D}$ is replaced with its finite subset in problem \eqref{1}. For each scenario $\bm{d}$ in the subset, we define a wait-and-see variable $\bm{y}_{\bm{d}}$, representing the action to be taken if scenario ${\bm{d}}$ is realized. The CCG algorithm falls into this class of methods.

\subsection{Column and Constraint Generation Algorithm}
\label{sec:CCG}
 The CCG algorithm is an iterative algorithm to solve problem \eqref{1} to near optimality. The first step of this algorithm is to reformulate the objective function as a function of here-and-now variables. Problem  \eqref{1} can be 
 reformulated as  
\begin{equation}
\underset{\bm{x}}\min\quad \Bigg(\max_{\bm{d}\in\mathcal{D}} \min_{\bm{y}\in \Omega(\bm{x},\bm{d})} \bm{c}(\bm{d})^\top \bm{x} + \bm{b}^\top \bm{y}\Bigg),
\label{2}
\end{equation}
with $\Omega(\bm{x},\bm{d}) = \{\bm{y}: \bm{A}(\bm{d})\bm{x} + \bm{By} \leq \bm{g} \}$.
We also define  
\begin{align}
\label{3}
\mathcal{Q}(\bm{x}) = \max_{\bm{d}\in\mathcal{D}} \min_{\bm{y}_{\bm{d}}\in \Omega(\bm{x},\bm{d})} \bm{c}(\bm{d})^\top \bm{x} + \bm{b}^\top \bm{y}_{\bm{d}},
\end{align}
which is the objective value corresponding to a here-and-now decision $\bm{x}$. The solution to the outer maximization problem in \eqref{3} is the worst-case scenario associated with $\bm{x}$. 

Since a worst-case scenario is an extreme point of the uncertainty set, in the inner maximization of problem \eqref{2}, it suffices to optimize over the extreme points of $\mathcal{D}$ rather than the entire set. Hence, problem \eqref{2} is equivalent to the following problem.
\begin{alignat}{2}
&\underset{\bm{x}, \bm{\alpha}, \{\bm{y_{\bm{d}}}\}_{\bm{d} \in \mathcal{E}}} \min\quad && \alpha \label{4} \\
&\text{s.t.} && \alpha \geq \bm{c}(\bm{d})^\top \bm{x} + \bm{b}^\top \bm{y}_{\bm{d}} , \quad \forall  \bm{d} \in \mathcal{E}, \nonumber \\
&  && \bm{y}_{\bm{d}} \in \Omega(\bm{x},\bm{d}), \quad \forall  \bm{d} \in \mathcal{E}, \nonumber 
\end{alignat}
\noindent where $\mathcal{E}$ is the set of all extreme points of $\mathcal{D}$ and $\bm{y}_{\bm{d}}$ is the wait-and-see variable associated with $\bm{d}$. 

Without using the entire set $\mathcal{E}$, CCG solves \eqref{4} by iteratively adding a new extreme point $\bm{d}$ and the  associated wait-and-see variable $\bm{y}_{\bm{d}}$ until a convergence criterion is met. {\color{black} In the initial iteration, where no extreme point is identified yet ($\mathcal{E}_0 = \emptyset$), we solve \eqref{3}  with any  initial here-and-now decision $\bm{x}_0$ to find the associated worst-case scenario and then add this scenario to $\mathcal{E}_0$.} Iteration $i$ of the CCG algorithm involves $i$ extreme points identified so far. Denoting the set of extreme points at iteration $i$ as $\mathcal{E}_i$, the so-called restricted master problem at iteration $i$ is
\begin{alignat}{2}
&\underset{\bm{x}, \bm{\alpha}, \{\bm{y}_{\bm{d}}\}_{{\bm{d} \in \mathcal{E}_i}}}\min\quad &&\alpha  \label{5}\\
&\text{s.t.} && \alpha \geq \bm{c}(\bm{d})^\top \bm{x} + \bm{b}^\top \bm{y}_{\bm{d}}, \quad \forall  {\bm{d} \in\ \mathcal{E}_i}, \nonumber\\
&\qquad \ && {\bm{y}_{\bm{d}} \in \Omega(\bm{x},\bm{d})}, \quad \forall {\bm{d} \in\ \mathcal{E}_i}.  \nonumber
\end{alignat}
The objective value of \eqref{5} is lower than the objective value of \eqref{4}, as only a subset of the constraints are imposed. Once we solve \eqref{5} and obtain its solution $\bm{x}_i$, we calculate $\mathcal{Q}(\bm{x}_i)$ and also obtain the worst-case scenario $\bm{d}_i$ associated with $\bm{x}_i$. If the gap between $\mathcal{Q}(\bm{x}_i)$ and the objective value of \eqref{5} is larger than a convergence criterion, $\bm{d}_i$ is added to $\mathcal{E}_i$ from the next iteration. 

In every iteration, we have to evaluate $\mathcal{Q}(\bm{x}_i)$, which is a non-convex max-min problem as shown in \eqref{3}. Several methods have been proposed to solve this problem, such as converting it to a mixed integer linear optimization problem \citep{CCG,Sun2014Daily}. In our implementation, we use a heuristic called the Alternating Direction method due to its computational efficiency and strong empirical performance. 
%\textcolor{black}{Specifically, \citep{Sun2014Daily} show that in each iteration, the exact method did not finish within 10 minutes, while the heuristic only takes 1.52 seconds. Nevertheless, the final objective cost obtained by the combination of CCG and the heuristic to be described are on average only 2\% different from the objective cost found by the combination of CCG and the exact method.} 

Using the strong duality in linear optimization, the inner minimization problem in \eqref{3} can be converted to a maximization problem. Now the problem \eqref{3} is recast into the following maximization problem. 
\begin{alignat}{2}
&\underset{\bm{d}, \bm{\pi}}\max\quad &&\bm{\pi}^\top (\bm{A}(\bm{d})\bm{x} - \bm{g}) + \bm{c}(\bm{d})^\top \bm{x}  \label{6} \\
&\text{s.t.} && -\bm{\pi}^\top \bm{B} = \bm{b}^\top, \nonumber\\
&\ && \bm{\pi} \geq \bm{0}, \nonumber\\
& \ && \bm{d} \in \mathcal{D}. \nonumber
\end{alignat}

% Note that the constraints on $\bm{d}$ and $\bm{\pi}$ are disjoint because of the fixed-recourse property. 
The Alternating Direction method to solve problem \eqref{6} is described in Algorithm \ref{alg2}. For conciseness, we use $\mathcal{T} = \{\bm{\pi} \mid -\bm{\pi}^\top \bm{B} = \bm{b}^\top, \bm{\pi} \geq 0\}$ in the algorithm description. Theoretically, CCG can output suboptimal solutions precisely because problem \eqref{3} is non-convex. Hence, in each iteration, we are in fact computing an approximation of $\mathcal{Q}(\bm{x}_i)$, which we denote as $\tilde{\mathcal{Q}}(\bm{x}_i)$. To ensure the quality of the solution, in our implementation we try three different initial points for problem \eqref{3} and choose the best solution found. For more detail on CCG method and the Alternating Direction method see \citep{Sun2014Wind, Sun2014Daily}. Algorithms \ref{alg1} and \ref{alg2} describe the CCG and the Alternating Direction method, respectively. 

\begin{algorithm}
\small
 \KwInput{Problem \eqref{1}, $\epsilon_1$}
\KwOutput{$\tilde{\bm{x}}^*$,$\tilde{\bm{d}}^*$} 
 \textbf{Initialization}: $i = 0$, $\bm{x}_0$, $\mathcal{E}_0 = \emptyset$,  UB = $\infty$, LB = $-\infty$\\
 \While{UB -LB $\geq \epsilon_1$}{

  \If{$i = 0$}{Evaluate $\mathcal{Q}(\bm{x}_i)$ to get $\tilde{\mathcal{Q}}(\bm{x}_i)$ and a solution $\bm{d}_i$ \\
  $\mathcal{E}_{i+1} \leftarrow \mathcal{E}_{i} \bigcup \{\bm{d}_i\}$\\
  $i$ $\leftarrow$ $i+1$
  }
  \Else{Solve \eqref{5} with the extreme points in $\mathcal{E}_i$. Denote the solutions as $\bm{x}_i$ and $\bm{\alpha}_i$. \\
  LB $\leftarrow$ $\bm{\alpha}_i$\\
  Evaluate $\mathcal{Q}(\bm{x}_i)$ to get $\tilde{\mathcal{Q}}(\bm{x}_i)$ and a solution $\bm{d}_i$.\\
  UB $\leftarrow$ $\tilde{\mathcal{Q}}(\bm{x}_i)$\\
  $\mathcal{E}_{i+1} \leftarrow \mathcal{E}_{i} \bigcup \{\bm{d}_i\}$\\
  $i$ $\leftarrow$ $i+1$}{
  
  }} 
  $\tilde{\bm{x}}^* \leftarrow \bm{x}_i$ \\
  $\tilde{\bm{d}}^* \leftarrow \bm{d}_i$
 \caption{Column and Constraint Generation}
 \label{alg1}
\end{algorithm}

\begin{algorithm}
\small
 \textbf{Input}: Problem \eqref{3}, $\bm{x}_i, \epsilon_2$ \\
\textbf{Output}: $\tilde{\mathcal{Q}}(\bm{x}_i)$, $\bm{d}_i$\\
 \textbf{Initialization}: Some $\bm{d}_0 \in\mathcal{D}, t = 0$, UB = $\infty$, LB = $-\infty$ \\
  
 \While{UB -LB $\geq \epsilon_2$}{
  LB $\leftarrow (a) \, \max_{\bm{\pi} \in \mathcal{T}} \bm{\pi}^\top (\bm{A}(\bm{d}_t)\bm{x}_i - \bm{g}) + \bm{c}(\bm{d}_t)^\top \bm{x}_i$ \\
  Denote the solution of (a) as $\bm{\pi}_{t}$.\\ 
  UB $\leftarrow (b) \, \max_{\bm{d} \in \mathcal{D}} \bm{\pi}_{t}^\top (\bm{A}(\bm{d})\bm{x}_i - \bm{g}) + \bm{c}(\bm{d})^\top \bm{x}_i$ \\
  Denote the solution of (b) as $\bm{d}_{t}$.\\
 $t$ $\leftarrow$ $t + 1$
  }
  $\tilde{\mathcal{Q}}(\bm{x}_i) \leftarrow \frac{\text{UB}+\text{LB}}{2}$ \\
  $\bm{d}_i \leftarrow \bm{d}_t$
 \caption{Alternating Direction Method}
 \label{alg2}
\end{algorithm}

\subsection{Obtaining the solutions}

Given problem \eqref{1} and a scenario $\bar{\bm{d}}$, we can find a near-optimal here-and-now decision $\tilde{\bm{x}}^*$  and an associated worst-case scenario $\tilde{\bm{d}}^*$ using Algorithm \ref{alg1}. We can find a near-optimal wait-and-see decision $\tilde{\bm{y}}^*(\bar{\bm{d}})$ by fixing $\bm{x} = \tilde{\bm{x}}^*$, $\bm{d} = \bar{\bm{d}}$ and solving the deterministic version of \eqref{1}, which is the following problem.
\begin{alignat*}{2}
&\underset{\bm{y}}\min\quad && \bm{c}(\bar{\bm{d}})^\top\tilde{\bm{x}}^{*} +\bm{b}^\top\bm{y}  \\ 
&s.t.\quad &&  \bm{A}(\bar{\bm{d}})\tilde{\bm{x}}^* + \bm{B}\bm{y} \leq \bm{g}.\quad \nonumber
\end{alignat*}

\noindent Note that the decision variable $\bm{y}$ is no longer a functional variable but a finite vector of decision variables, because the specific scenario $\bar{\bm{d}}$ has been realized.

\subsection{Scalability}
\label{sec:scale}

We demonstrate that solving ARO problems using Algorithm \ref{alg1} can be computationally demanding by considering the unit commitment problem from power systems literature. This problem involves minimizing energy production costs while satisfying energy demand over $m$ time steps for a power system with $n$ generators. We should decide which generators to start up, shut down, and how much energy each generator should produce at each time. Whether we should start up or shut down each generator at each time, referred to as the commitment decisions, are modeled as binary variables. In the ARO version, these variables represent the here-and-now decisions made before the demand is realized. The demand at each time is the uncertain parameter. After the demand is realized, the wait-and-see decisions determine how much energy each generator should produce. Supplementary Material provides the complete formulation of the deterministic version, and the data used is from \citep{UCdata}. We use the budget uncertainty set defined as $\mathcal{D} = \{\bm{d} \mid \sum_{i}\abs{\frac{d_i - \bar{d_i}}{0.1 \times \bar{d_i}}} \leq 2, \abs{d_i - \bar{d_i}} \leq 0.1 \times \bar{d_i}  \}$, where $\bar{\bm{d}}$ is from the data in \citep{UCdata}.  

\begin{figure}[h]
    \centering
    \includegraphics[width=0.4\textwidth]{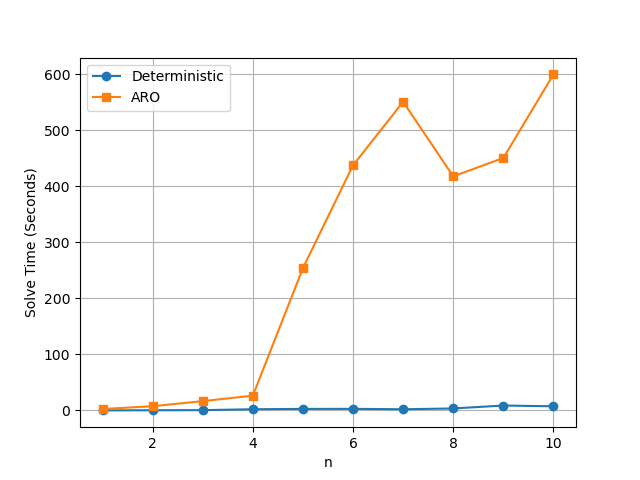}
    \caption{Solve time for the unit commitment problem.}
    \label{fig:comparion}
\end{figure}

{\color{black} For varying values of $n$, we compare the solve times of ARO and the deterministic version of the unit commitment problem. We keep $m = 24$ fixed for all experiments. For each $n$, we generate 100 deterministic instances and solve them with Gurobi \citep{gurobi} using the optimality gap of 0.01. Then, we solve the ARO versions of these problems using Algorithm \ref{alg1}, with tolerance set to $\epsilon_1 = 0.01$ for fair comparison. For Algorithm \ref{alg2} implemented within Algorithm \ref{alg1}, three random initial points are used and the tolerance is set to $\epsilon_2 = 0.01$. The experiment in this section was executed in Julia 1.4.1 on a MacBook Pro with 2.6 GHz Intel Core i7 CPU and 16GB of RAM. 

We present the experiment results in Figure \ref{fig:comparion}, where we report the mean solve times for each $n$. It is evident that compared to solving the deterministic version of the unit commitment problem, considerably more time is required to solve its ARO counterpart using Algorithm \ref{alg1}. The solve time for ARO increases drastically at $n = 5$, while the solve time for the deterministic version remains relatively consistent across all $n$ values. This finding suggests that solving ARO problems using Algorithm \ref{alg1} can be computationally challenging, especially for large scale problems.}

\section{A Machine Learning Approach to ARO}
\label{sec:Machine Learning for Adaptive Optimizaion}
In this section, we develop a machine learning approach to two-stage ARO with binary here-and-now variables and polyhedral uncertainty sets. In the context of MICO, \citet{Voice, Prune} use classification algorithms, while \citet{OPT} use a prescriptive machine learning algorithm, Optimal Policy Trees (OPT) \citep{Policy}. \citet{OPT} demonstrate that OPT has an edge over a classification algorithm, particularly in its ability to avoid infeasible or highly suboptimal solutions. In our work, we provide an integrated perspective by introducing both classification and prescriptive approaches in the context of ARO. {\color{black} We begin by presenting a comprehensive explanation of our foundational approach from Section \ref{subsec:strategy} to \ref{subsec:facility}. Following this, we introduce a generalization that extends the applicability of machine learning models to problems with varying dimensions in Section \ref{subsec:variability}.}

\subsection{Optimal Strategy}
\label{subsec:strategy}

We consider the following ARO problem, where $\bm{\theta}$ is the key parameter used to generate instances.
\begin{alignat}{2}
&\underset{\bm{x},\bm{y}(\cdot)}\min\quad && \Bigg(\max_{\bm{d}\in\mathcal{D}, }\bm{c}(\bm{d}, \bm{\theta})^\top\bm{x} +\bm{b}(\bm{\theta})^\top\bm{y}\Bigg) \label{aro} \\ 
&\text{s.t.}\quad &&  \bm{A}(\bm{d}, \bm{\theta})\bm{x} + \bm{B}(\bm{\theta})\bm{y} (\bm{d})\leq \bm{g}(\bm{\theta}),\quad \forall {\bm{d}\in\mathcal{D}}, \nonumber \\
& \quad && \bm{x} \; \text{is binary}. \nonumber
\end{alignat}
\noindent We denote the deterministic version of problem \eqref{aro} with fixed $\bm{x} = \bm{x}^*, \bm{d} = \bm{d}^*$ as $Det(\bm{\theta}, \bm{x}^*,\bm{d}^*)$, which is the following problem.
\begin{alignat*}{2}
&\underset{\bm{y}}\min\quad && \bm{c}(\bm{d}^*, \bm{\theta})^\top\bm{x}^{*} +\bm{b}(\bm{\theta})^\top\bm{y}  \quad \tag{$Det(\bm{\theta}, \bm{x}^*,\bm{d}^*)$}\\ 
&\text{s.t.}\quad &&  \bm{A}(\bm{d}^*, \bm{\theta})\bm{x}^* + \bm{B}(\bm{\theta})\bm{y} \leq \bm{g}(\bm{\theta}).\quad \nonumber
\end{alignat*}

\noindent We denote the optimal objective cost of $Det(\bm{\theta}, \bm{x}^*,\bm{d}^*)$ as $V(\bm{\theta}, \bm{x}^*,\bm{d}^*)$ and assume $Det(\bm{\theta}, \bm{x}^*,\bm{d}^*)$ has $m$ constraints.

Given the ARO instance \eqref{aro} with a parameter $\bar{\bm{\theta}}$ and a scenario $\bar{\bm{d}} \in \mathcal{D}$, we define the optimal strategy for the here-and-now decisions, the worst-case scenarios associated with the optimal here-and-now decisions, and the wait-and-see decisions associated with the scenario $\bar{\bm{d}}$. We denote these strategies as $s_{\bm{x}}(\bar{\bm{\theta}}), s_{\bm{d}}(\bar{\bm{\theta}}), s_{\bm{y}}(\bar{\bm{\theta}}, \bar{\bm{d}})$, respectively. We use $s_{\bm{x}}, s_{\bm{d}}, s_{\bm{y}}$ instead when we are not referring to a specific instance or scenario. These strategies serve as the prediction targets for the trained machine learning models. In the following description, we use $\bm{x}^*, \bm{d}^*$ to denote the optimal here-and-now decision and the worst-case scenario associated with the optimal here-and-now decision, respectively. 

% \begin{figure}
%     \centering
%     \includegraphics[scale = 0.3]{image/Objective.png}
%     \caption{Overview of the Objectives.}
%     \label{fig:objective}
% \end{figure}

\paragraph{Here-and-now Decisions}
The here-and-now variables in problem \eqref{aro} are binary. Therefore, we define the optimal strategy for the here-and-now decisions as the optimal here-and-now decision itself, meaning that $s_{\bm{x}}(\bar{\bm{\theta}}) = \bm{x}^*$. Once a machine learning model is trained, it can directly predict a here-and-now decision given a new parameter $\hat{\bm{\theta}}$.

\paragraph{Worst-case Scenarios}

%the exact point can be recovered by imposing the tight constraints as equalities. Thus, the information necessary to acquire the worst-case scenario can be summarized in $(\bm{x}^*, \tau_d(\bm{x}^*))$, where $\tau_d$ denotes the tight constraints of the constraints $\bm{d}^{*} \in \mathcal{D}$. Here $\tau_d(\bm{x}^*)$ is used to stress the fact that the worst-case scenario depends on the here-and-now decision. 

 In general, it is hard to define a single worst-case scenario of an ARO instance. However, if we fix some here-and-now decision, the worst-case scenario that corresponds to this specific decision can be defined. We define the optimal strategy for the worst-case scenarios as  $s_{\bm{d}}(\bar{\bm{\theta}}) =(\bm{x}^*,\bm{d}^*)$. Note that since a worst-case scenario is one of the extreme points of $\mathcal{D}$, there can only be a finite number of worst-case scenarios possible. Once a machine learning model is trained, it can directly predict a here-and-now decision and a worst-case scenario given a new parameter $\hat{\bm{\theta}}$.

\paragraph{Wait-and-see Decisions}
Given $\bm{x}^*$ and the scenario $\bar{\bm{d}}$,  we solve $Det(\bar{\bm{\theta}},\bm{x}^*, \bar{\bm{d}})$ to identify the optimal solution $\bm{y}^*$ and the set of constraints that are satisfied as equality at optimality. These constraints are referred to as the tight constraints and are denoted as $\tau_{\bm{y}}(\bar{\bm{\theta}},\bar{\bm{d}})$. Formally, they are defined as
\[ \tau_{\bm{y}}(\bar{\bm{\theta}},\bar{\bm{d}}) = \{j \in [m] \mid [\bm{A}(\bar{\bm{d}}, \bar{\bm{\theta}})\bm{x}^* + \bm{B}(\bar{\bm{\theta}})\bm{y}^*]_j = [\bm{g}(\bar{\bm{\theta}})]_j \}. \]

\noindent Identifying the tight constraints simplifies the linear programming problem, as unnecessary constraints can be removed. We define the optimal strategy as $s_{\bm{y}}(\bar{\bm{\theta}}, \bar{\bm{d}}) = (\bm{x}^*, \tau_{\bm{y}}(\bar{\bm{\theta}},\bar{\bm{d}}))$. Given a new parameter $\hat{\bm{\theta}}$ and a scenario $\hat{\bm{d}}$, a model predicts a here-and-now decision $\hat{\bm{x}}$ and a set of tight constraints. A wait-and-see decision for the scenario $\hat{\bm{d}}$ can be computed by solving $Det(\hat{\bm{\theta}}, \hat{\bm{x}}, \hat{\bm{d}})$ imposing only the predicted tight constraints.

Notice that the optimal strategies for the worst-case-scenarios and the wait-and-see decisions already contain the optimal here-and-now decisions. Therefore, it might seem redundant to train a separate model to predict the optimal here-and-now decisions. However, we demonstrate in Section \ref{sec:numerical experiment} that the prediction accuracy for the here-and-now decisions is generally higher than the other prediction targets. Hence, if one is only interested in predicting the optimal here-and-now decisions, training a separate model might be beneficial.

\subsection{Suboptimality and Infeasibility}
\label{subsec:measure}

We explain how we evaluate the quality of the strategies applied to an ARO instance associated with $\hat{\bm{\theta}}$ and a scenario $\hat{\bm{d}}$. We denote the strategies that we would like to evaluate as $\hat{\bm{x}}, (\hat{\bm{x}}, {\hat{\bm{d}}^{'}}), (\hat{\bm{x}}, {\hat{\tau}_{\bm{y}}})$, respectively.

\paragraph{Here-and-now Decisions}

Suboptimality and infeasibility of the strategy $\hat{\bm{x}}$ are measured by comparing $\tilde{\mathcal{Q}}(\hat{\bm{x}})$ and $\tilde{\mathcal{Q}}(\tilde{\bm{x}}^*)$ of the instance associated with $\hat{\bm{\theta}}$. We consider $\hat{\bm{x}}$ infeasible if $\tilde{\mathcal{Q}}(\hat{\bm{x}}) = \infty$. If it is feasible, we define its suboptimality as 
\[sub(\hat{\bm{x}}) = (\tilde{\mathcal{Q}}(\hat{\bm{x}}) - \tilde{\mathcal{Q}}(\tilde{\bm{x}}^*)) / \abs{\tilde{\mathcal{Q}}(\tilde{\bm{x}}^*)}.\]

\paragraph{Worst-Case Scenarios}
Measuring the quality of the strategy $(\hat{\bm{x}}, \hat{\bm{d}}^{'})$ consists of two stages. First, we evaluate the $\hat{\bm{x}}$ part following the procedure described above for the here-and-now decisions. If $\hat{\bm{x}}$ is infeasible, the strategy $(\hat{\bm{x}}, \hat{\bm{d}}^{'})$ is considered infeasible in the first place. Otherwise, it is considered feasible. If it is feasible, then now we check if $\hat{\bm{d}}^{'}$ is the worst-case scenario for $\hat{\bm{x}}$. To do so, we solve $Det(\hat{\bm{\theta}}, \hat{\bm{x}},\hat{\bm{d}}^{'})$ and compare the optimal cost with $\tilde{\mathcal{Q}}(\hat{\bm{x}})$. We define the suboptimality as

\[sub(\hat{\bm{x}}, \hat{\bm{d}}^{'}) = \max  \Bigg\{ sub(\hat{\bm{x}}), \bigg(\tilde{\mathcal{Q}}(\hat{\bm{x}}) -  V(\hat{\bm{\theta}}, \hat{\bm{x}},\hat{\bm{d}}^{'})  \bigg) / \abs{\tilde{\mathcal{Q}}(\hat{\bm{x}})} \Bigg\}.\]

% Note that the evaluation criteria for the worst-case scenarios is more strict than the criteria for the here-and-now decisions. When the first component of a prediction $(\hat{\bm{x}}, \hat{\bm{d}})$, $\hat{\bm{x}}$, is merely suboptimal, we simply regard it infeasible. This is to focus our attention on evaluating the worst-case scenario part of the strategy, rather than the here-and-now decision part. That is, we would like to see whether $\hat{\bm{d}}$ is the worst-case scenario corresponding to $\hat{\bm{x}}$, rather than gauging the quality of the $\hat{\bm{x}}$ part itself. 

\paragraph{Wait-and-see Decisions}

Measuring the quality of the strategy $(\hat{\bm{x}}, \hat{\tau}_{\bm{y}})$ also consists of two stages. First, we evaluate the $\hat{\bm{x}}$ part following the procedure described above for the here-and-now decisions. If it is infeasible, then $(\hat{\bm{x}}, \hat{\tau}_{\bm{y}})$ is considered infeasible in the first place. If it is feasible, then we evaluate the $\hat{\tau}_{\bm{y}}$ part. We solve $Det(\hat{\bm{\theta}}, \hat{\bm{x}}, \hat{\bm{d}})$ imposing only the constraints included in $\hat{\tau}_{\bm{y}}$. If this leads to infeasibility, then again $(\hat{\bm{x}}, \hat{\tau}_{\bm{y}})$ is considered infeasible. If a feasible solution $\hat{\bm{y}}$ is found, we compute the suboptimality of the $\hat{\tau}_{\bm{y}}$ part defined as
 \begin{equation*}
sub(\hat{\tau}_{\bm{y}}) = \Bigg(\Big({\bm{c}(\hat{\bm{d}}, \hat{\bm{\theta}})^\top\hat{\bm{x}} +\bm{b}(\hat{\bm{\theta}})^\top\hat{\bm{y}}} \Big) - V(\hat{\bm{\theta}}, \hat{\bm{x}}, \hat{\bm{d}}) \Bigg) / V(\hat{\bm{\theta}}, \hat{\bm{x}}, \hat{\bm{d}}),
\end{equation*} 
For a feasible $(\hat{\bm{x}}, \hat{\tau}_{\bm{y}})$, we define its suboptimality as
\[sub(\hat{\bm{x}}, \hat{\tau}_{\bm{y}}) = \max  \bigg\{ sub(\hat{\bm{x}}),  sub(\hat{\tau}_{\bm{y}}) \bigg\}.\]

\subsection{A Classification Approach}
\label{sec:Algorithm}

In this section, we develop an approach to solve ARO problems using classification algorithms. The proposed approach consists of three phases. In Phase 1, we generate $N \in \mathbb{N} $ parameters $\{\bm{\theta}_i\}_{i \in [N]}$ and solve the associated ARO instances using Algorithm \ref{alg1}. For each $\bm{\theta}_i$, we also sample a scenario $\bm{d}_i$ from the uncertainty set. Then, we identify near-optimal or slightly suboptimal strategies for the here-and-now decisions, the worst-case scenarios associated with the here-and-now decisions, and the wait-and-see decisions associated with the scenario $\bm{d}_i$. In Phase 2, we use classification algorithms to train three machine learning models that predict each of these strategies. In Phase 3, given a new parameter $\hat{\bm{\theta}}$ and a scenario $\hat{\bm{d}}$, we use the predictions of these models to compute a here-and-now decision, a worst-case scenario, and a wait-and-see decision associated with the scenario $\hat{\bm{d}}$.  Algorithm \ref{alg:overview} provides a comprehensive overview of the entire procedure with detailed steps.

\paragraph{Remark 1} As mentioned in Section \ref{sec:CCG}, solving ARO problems to exact optimality is hard in general. Hence, we use the outputs of Algorithm \ref{alg1} to compute near-optimal or slightly suboptimal strategies instead in Phase 1.  The suboptimalities of the strategies depend on the tolerance $\epsilon_1$ in Algorithm \ref{alg1}.

\paragraph{Remark 2} In this work, we sample parameters $\bm{\theta}_i, i \in [N]$, uniformly at random from the ball with a predefined radius $r$. We sample scenarios $\bm{d}_i$ from the uncertainty set, also uniformly at random. Depending on the application area or any prior knowledge, the sampling scheme may vary.

\begin{algorithm}[ht]
\small
 \textbf{Input}: $\bar{\bm{\theta}}$, $N$, $r$, Problem \eqref{aro}, $\epsilon_1$, $\epsilon_2$ 
 \vspace{2mm}
 
 \textbf{Phase 1}\\
 1.1 \For{$i = 1, \dots, N$}{
  Sample a point $\bm{\theta}_i$ from the ball $\mathcal{B}(\bar{\bm{\theta}}, r)$ uniformly at random. \\
  Fix $ \bm{\theta} = \bm{\theta}_i$ and solve problem \eqref{aro} with Algorithm \ref{alg1} to obtain a here-and-now decision $\tilde{\bm{x}}^*_i$ and an associated worst-case-scenario $\tilde{\bm{d}}^*_i$. The tolerances for Algorithm \ref{alg1} and \ref{alg2} are set to $\epsilon_1$ and $\epsilon_2$, respectively. \\
  Sample a point $\bm{d}_i$ from $\mathcal{D}$ uniformly at random.\\
  Solve $Det(\bm{\theta}_i, \tilde{\bm{x}}^*_i, \bm{d}_i)$ to obtain $\tilde{\tau}_{\bm{y}}(\bm{\theta}_i, \bm{d}_i)$. \\
  $s_{\bm{x}}(\bm{\theta}_i) \leftarrow \tilde{\bm{x}}^*_i$\\
  $s_{\bm{d}}(\bm{\theta}_i) \leftarrow (\tilde{\bm{x}}^*_i, \tilde{\bm{d}}^*_i)$ \\
  $s_{\bm{y}}(\bm{\theta}_i, \bm{d}_i) \leftarrow (\tilde{\bm{x}}^*_i, \tilde{\tau}_{\bm{y}}(\bm{\theta}_i, \bm{d}_i)) $ \\
  }

     \vspace{2mm}

\textbf{Phase 2}\\
  2.1 Train a machine learning model $\mathcal{L}_{\bm{x}}$ using $(\bm{\theta}_1, \dots, \bm{\theta}_N)$ as the feature matrix and $\Big(s_{\bm{x}}(\bm{\theta}_1), \dots , s_{\bm{x}}(\bm{\theta}_N) \Big)$ as the target vector.\\
  2.2 Train a machine learning model $\mathcal{L}_{\bm{d}}$ using $(\bm{\theta}_1, \dots, \bm{\theta}_N)$ as the feature matrix and $\Big(s_{\bm{d}}(\bm{\theta}_1), \dots , s_{\bm{d}}(\bm{\theta}_N) \Big)$ as the target vector.\\
  2.3 Train a machine learning model $\mathcal{L}_{\bm{y}}$ using $\Big((\bm{\theta}_1, \bm{d}_1), \dots, (\bm{\theta}_N, \bm{d}_N) \Big)$ as the feature matrix and  $\Big(s_{\bm{y}}(\bm{\theta}_1, \bm{d}_1), \dots , s_{\bm{y}}(\bm{\theta}_N, \bm{d}_N) \Big)$ as the target vector. \\
  
  \vspace{2mm}

  \textbf{Phase 3}\\
  3.1 For a new instance $\hat{\bm{\theta}}$, $\mathcal{L}_{\bm{x}}$ predicts $\hat{s}_{\bm{x}}(\hat{\bm{\theta}}) = \hat{\bm{x}}$. \\
  3.2 For a new instance $\hat{\bm{\theta}}$, $\mathcal{L}_{\bm{d}}$ predicts $\hat{s}_{\bm{d}}(\hat{\bm{\theta}}) = (\hat{\bm{x}},\hat{\bm{d}}^{'})$. \\ 
  3.3.1 For a new instance $\hat{\bm{\theta}}$ and  $\hat{\bm{d}}$, $\mathcal{L}_{\bm{y}}$ predicts $\hat{s}_{\bm{y}}(\hat{\bm{\theta}}, \hat{\bm{d}}) = (\hat{\bm{x}}, \hat{\tau}_{\bm{y}}(\hat{\bm{\theta}}, \hat{\bm{d}}))$.\\
  3.3.2 Solve $Det(\hat{\bm{\theta}},\hat{\bm{x}}, \hat{\bm{d}}) $ using $\hat{\tau}_{\bm{y}}(\hat{\bm{\theta}}, \hat{\bm{d}}) $ to compute a wait-and-see decision.
  
 \caption{Classification Approach to ARO}
 \label{alg:overview}
\end{algorithm}

% \paragraph{Remark} Our approach can be extended to two-stage ARO with general integer here-and-now variables and polyhedral uncertainty sets. In this paper, we focus on the problems with binary here-and-now variables.

\subsection{A Prescriptive Approach}
\label{sec:prescriptive}

In this section, we present a prescriptive approach to ARO using OPT. We begin with a baseline approach and then generalize it. In the following explanation, we focus on training a machine learning model for the here-and-now variables, but the cases of worst-case scenarios and the wait-and-see decisions are straightforward extensions.

The baseline approach is similar to the classification approach provided in Algorithm \ref{alg:overview}, except for Phase 2. In Phase 2 of the prescriptive approach, we need to compute what we refer to as the reward matrices. 

Assume that after Phase 1 of Algorithm \ref{alg:overview}, we have solved $N \in \mathbb{N}$ ARO instances and identified $Q\in \mathbb{N}$ different strategies in the training set. Note that $Q \leq N$, as the optimal strategies of different instances might overlap. We let $\mathcal{S}_{\bm{x}} = \{s_{{\bm{x}},1}, \dots, s_{{\bm{x}},Q}\}$ be the set of optimal strategies identified, where $s_{{\bm{x}},i} \neq s_{{\bm{x}},j}$ if $i \neq j$.  The reward matrix $\bm{R_x} \in \mathbb{R}^{N \times Q}$ is then defined such that its entry in the $i_{th}$ row and $j_{th}$ column corresponds to the suboptimality of the strategy $s_{{\bm{x}},j}$ applied to the $i_{th}$ ARO instance. If the strategy is infeasible, we assign an arbitrary large number to its entry. Using the reward matrix, we train a decision tree by solving the optimization problem
\begin{align*} 
&\underset{v(\cdot),\bm{z}} \min \quad  \sum_{i=1}^{N}\sum_{\ell}\mathbbm{1}\{v(\bm{\theta}_i) = \ell\}{\cdot}R_{iz_{\ell}},
\end{align*}
\noindent where $v(\bm{\theta}_i)$ is the leaf of the tree $\bm{\theta}_i$ is assigned to, $\bm{z}_{\ell}$ is the strategy assigned to the points in leaf $\ell$, and $R_{iz_{\ell}}$ is the suboptimality of the instance $i$ under the strategy $z_{\ell}$. This optimization problem determines the structure of the decision tree using the decision variable $v(\cdot)$ and assigns strategies to each leaf using the decision variable $\bm{z}$. The objective is to train a decision tree that prescribes a strategy to an ARO instance, so that the resulting suboptimality is minimized. Once a decision tree is trained, given a feature vector $\hat{\bm{\theta}}$, we traverse the tree using the feature until we reach the leaf node. The strategy assigned to this leaf is the prediction for the instance $\hat{\bm{\theta}}$. For a more detailed explanation, please refer to \citep{Policy}.

Now, we introduce a generalization of the baseline approach just explained. In our computational experiments, we have observed that the number $Q$ can get prohibitively large, especially for large scale problems. While we propose a method to address this issue when training a model for the wait-and-see decisions in Section \ref{sec5}, this method does not extend to other prediction targets.

In the generalization we propose, we randomly select $Q_1 \leq Q$ strategies from $\mathcal{S}_{\bm{x}}$, and compute the corresponding reward matrix. Using this reward matrix, we train a decision tree. Algorithm \ref{alg:prescriptive} outlines the entire procedure.

\begin{algorithm}[ht]
\small
 \textbf{Input}: $\bar{\bm{\theta}}$, $N$, $r$, Problem \eqref{aro}, $\epsilon_1$, $\epsilon_2$, $M, Q_1, Q_2, Q_3$ \
 
 \vspace{2mm}
 
 \textbf{Phase 1}\\
 1.1 Identical to 1.1 of Algorithm \ref{alg:overview}. \\
  1.2 $\mathcal{S}_{\bm{x}} \leftarrow \Big\{s_{\bm{x}}(\bm{\theta}_1), \dots , s_{\bm{x}}(\bm{\theta}_N) \Big\}$\\
     $\quad \ \mathcal{S}_{\bm{d}} \leftarrow \Big\{s_{\bm{d}}(\bm{\theta}_1), \dots , s_{\bm{d}}(\bm{\theta}_N) \Big\}$\\
     $\quad \ \mathcal{S}_{\bm{y}} \leftarrow \Big\{s_{\bm{y}}(\bm{\theta}_1, \bm{d}_1), \dots , s_{\bm{y}}(\bm{\theta}_N, \bm{d}_N) \Big\}$\\

     \vspace{2mm}

\textbf{Phase 2}\\
2.1.1 Choose $Q_1$ distinct strategies from $\mathcal{S}_{\bm{x}}$ at random and compute the reward matrix $\bm{R}_{\bm{x}} \in \mathbb{R}^{N \times Q_1}$ using those strategies. \\
  2.1.2 Train a decision tree $\mathcal{T}_{\bm{x}}$ using $\bm{R}_{\bm{x}}$. \\
2.2.1 Choose $Q_2$ distinct strategies from $\mathcal{S}_{\bm{d}}$ at random and compute the reward matrix $\bm{R}_{\bm{d}} \in \mathbb{R}^{N \times Q_2}$ using those strategies. \\
  2.2.2 Train a decision tree $\mathcal{T}_{\bm{d}}$ using $\bm{R}_{\bm{d}}$. \\
2.3.1 Choose $Q_3$ distinct strategies from $\mathcal{S}_{\bm{y}}$ at random and compute the reward matrix $\bm{R}_{\bm{y}} \in \mathbb{R}^{N \times Q_3}$ using those strategies. \\
  2.3.2 Train a decision tree $\mathcal{T}_{\bm{y}}$ using $\bm{R}_{\bm{y}}$. \\
  
  \vspace{2mm}

  \textbf{Phase 3}\\
  3.1 For a new instance $\hat{\bm{\theta}}$, $\mathcal{T}_{\bm{x}}$ predicts $\hat{s}_{\bm{x}}(\hat{\bm{\theta}}) = \hat{\bm{x}}$. \\
  3.2 For a new instance $\hat{\bm{\theta}}$, $\mathcal{T}_{\bm{d}}$ predicts $\hat{s}_{\bm{d}}(\hat{\bm{\theta}}) = (\hat{\bm{x}},\hat{\bm{d}}^{'})$. \\ 
  3.3.1 For a new instance $\hat{\bm{\theta}}$ and  $\hat{\bm{d}}$, $\mathcal{T}_{\bm{y}}$ predicts $\hat{s}_{\bm{y}}(\hat{\bm{\theta}}, \hat{\bm{d}}) = (\hat{\bm{x}}, \hat{\tau}_{\bm{y}}(\hat{\bm{\theta}}, \hat{\bm{d}}))$.\\
  3.3.2 Solve $Det(\hat{\bm{\theta}},\hat{\bm{x}}, \hat{\bm{d}}) $ using $\hat{\tau}_{\bm{y}}(\hat{\bm{\theta}}, \hat{\bm{d}}) $ to compute a wait-and-see decision.
  
 \caption{OPT for ARO}
 \label{alg:prescriptive}
\end{algorithm}

\subsection{Example}
\label{subsec:facility}

In this section, we apply Algorithm \ref{alg:prescriptive} to a small sized example and present the actual decision trees learned with OPT. We consider the following facility location problem formulated as an ARO problem.
%Policy  trees  are    learners  which assign a feature vector to an optimal prescription that optimizes an outcome. 
%Instead of predicting the optimal strategy, we consider the objective as prescribing it so that the objective value, which corresponds to the outcome in prescriptive settings, is minimized. Parameter tuning process is  skipped and the maximum depth of the trees is suppressed to 2 for simplicity. We utilize the interpretability of trees and convey intuition on the rationals behind some of the splits. Consider the following facility location problem. 
\begin{alignat}{3}
&\underset{\bm{y}(\cdot),\bm{x}}\min \; \underset{\bm{d} \in \mathcal{D}}\max \quad && \sum_{i =1}^{n} \sum_{j=1}^{m}c_{ij}y_{ij}(\bm{d}) + \sum_{i =1}^{n} f_ix_i \nonumber\\
&\text{s.t.}\quad && \sum_{i =1}^{n} y_{ij}(\bm{d}) \geq d_j, \quad &&\forall {\bm{d}\in\mathcal{D}}, \; \forall {j\in [m]}, \nonumber \\
&\quad && \sum_{j =1 }^{m} y_{ij}(\bm{d}) \leq p_ix_i, \quad && \forall {\bm{d}\in\mathcal{D}}, \;\forall {i\in[n]}, \nonumber \\
&\quad && y_{ij}(\bm{d}) \geq 0, \quad && \forall {\bm{d}\in\mathcal{D}}, \; \forall {i\in {[n]}}, \; \forall {j\in [m]}, \nonumber \\
&\quad && \bm{x} \in \{0,1\}^{n}. \nonumber
\end{alignat}

 Let $i \in [n]$ denote a possible location to build facilities, and $j \in [m]$ denote a delivery destination. Let $c_{ij}$ be the cost of transporting goods from location $i$ to destination $j$ and $p_i$ be the capacity of the facility built on location $i$. The construction cost to build a facility on location $i$ is denoted as $f_i$. The demand at destination $j$ is denoted as $d_j$, which is the uncertain parameter. The demand is assumed to be realized after the construction decisions and before the delivery decisions are made. Binary variable $x_i$ is the here-and-now variable representing whether we build facility on location $i$ or not. $y_{ij}$ is the amount of goods to transport from $i$ to $j$, which is the wait-and-see variable. 
 
 The parameter we use to generate instances is the coefficient vector $\bm{c}$ of the cost function. The vector $\bm{c}$ is sampled uniformly from the ball $B(\bar{\bm{c}}, 1)$, where each entry of  $\bar{\bm{c}}$ is drawn from $U(2,4)$. Capacities $p_i$ are sampled from $U(8, 18)$ and $f_i$ is sampled from $U(3,5)$. The uncertainty set is defined as $\mathcal{D} = \bigg\{\bm{d} \mid \sum_{i}d_i \leq 16, 4 \leq  d_i \leq 6  \bigg\}$. 
 
 For the purpose of illustration, we use a small sized example with $n = m = 3$.  We set  $Q_1 = |\mathcal{S}_{\bm{x}}| , Q_2 = |\mathcal{S}_{\bm{d}}|$ and $ Q_3 = |\mathcal{S}_{\bm{y}}|$. In other words, the entire set of strategies found is used for training. The tolerances are set to $\epsilon_1 = \epsilon_2 = 0.001$, and the penalty for infeasible predictions are set to $M = 1000000$. We limit the maximum depth of the tree to two in order to develop intuition on the learned models. Furthermore, we assign a number to each constraint in the deterministic version of the problem to clarify which constraint we are referring to in the following description. The demand satisfaction constraint at destination $j, j \in [3]$, is denoted constraint $j$. The capacity constraint at location $i, i \in [3]$, is denoted constraint $i+3$. The non-negativity constraint on the amount of goods to transport from $i$ to $j$ is denoted constraint $6 + 3(i-1) + j$.

% \begin{figure}
%     \centering
%     \includegraphics[scale = 0.3]{image/here.png}
%     \caption{Decision tree to predict the optimal strategies for the here-and-now decisions.}
%     \label{fig:here}
% \end{figure}

\begin{figure}[htp]% [H] is so declass\'e!
\centering
\begin{minipage}{0.45\textwidth}
\includegraphics[width=\textwidth]{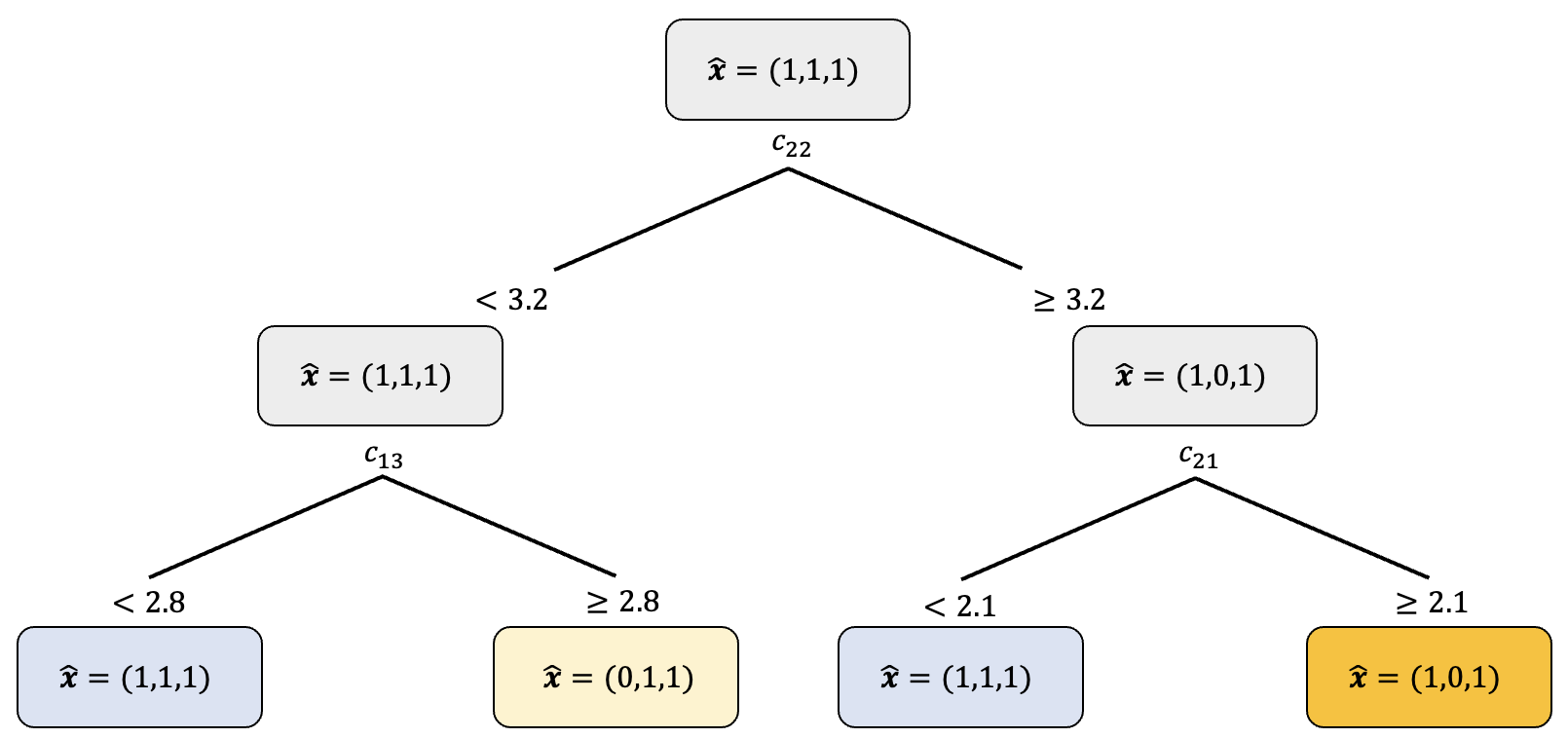}
\caption{Decision tree to predict the optimal strategies for the here-and-now decisions.}
\label{fig:here}
\end{minipage}\hfill
\begin{minipage}{0.45\textwidth}
\includegraphics[width=\textwidth]{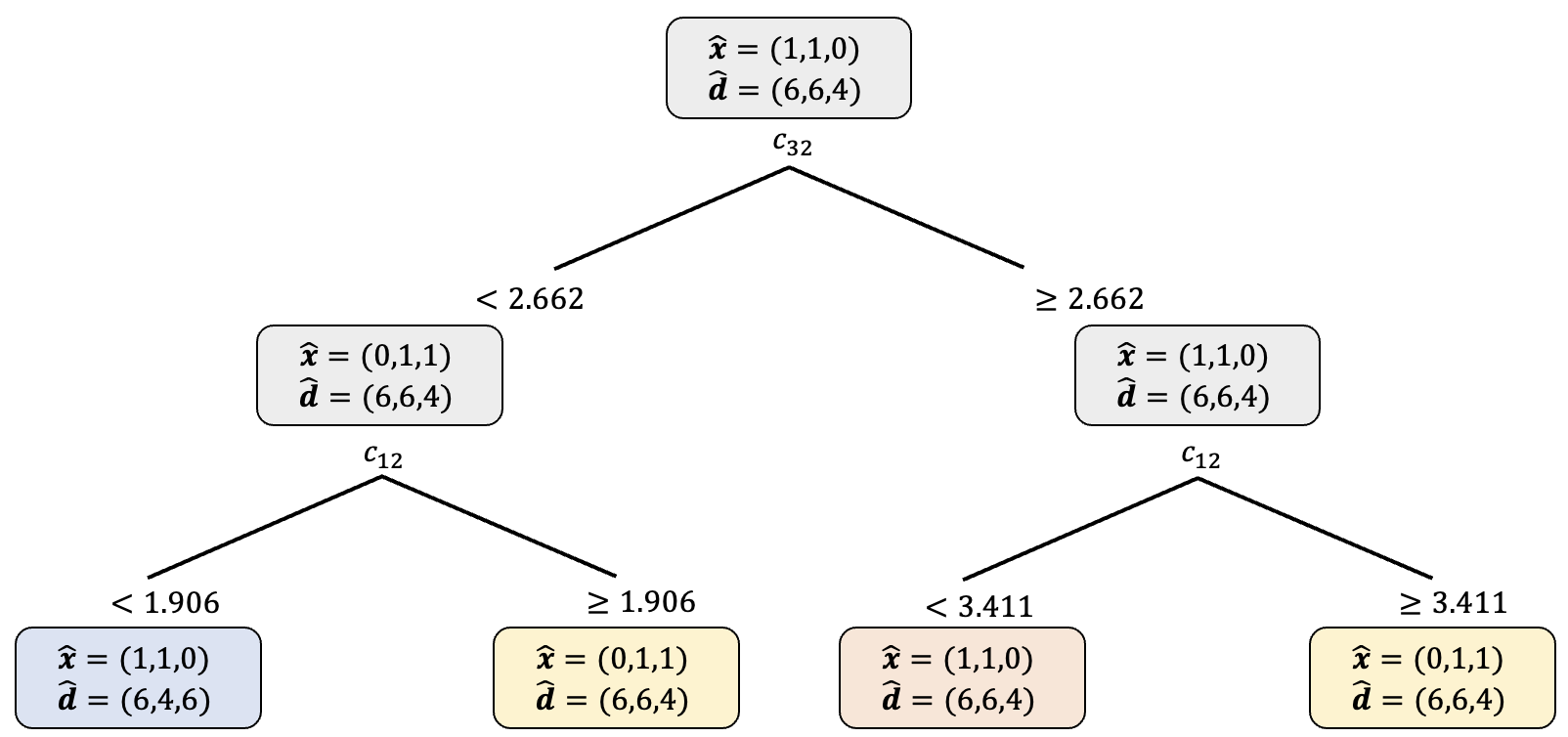}
\caption{Decision tree to predict the optimal strategies for the worst-case scenarios.}
\label{fig:scenario}
\end{minipage}\par
\vskip\floatsep% normal separation between figures
\includegraphics[width=0.7\textwidth]{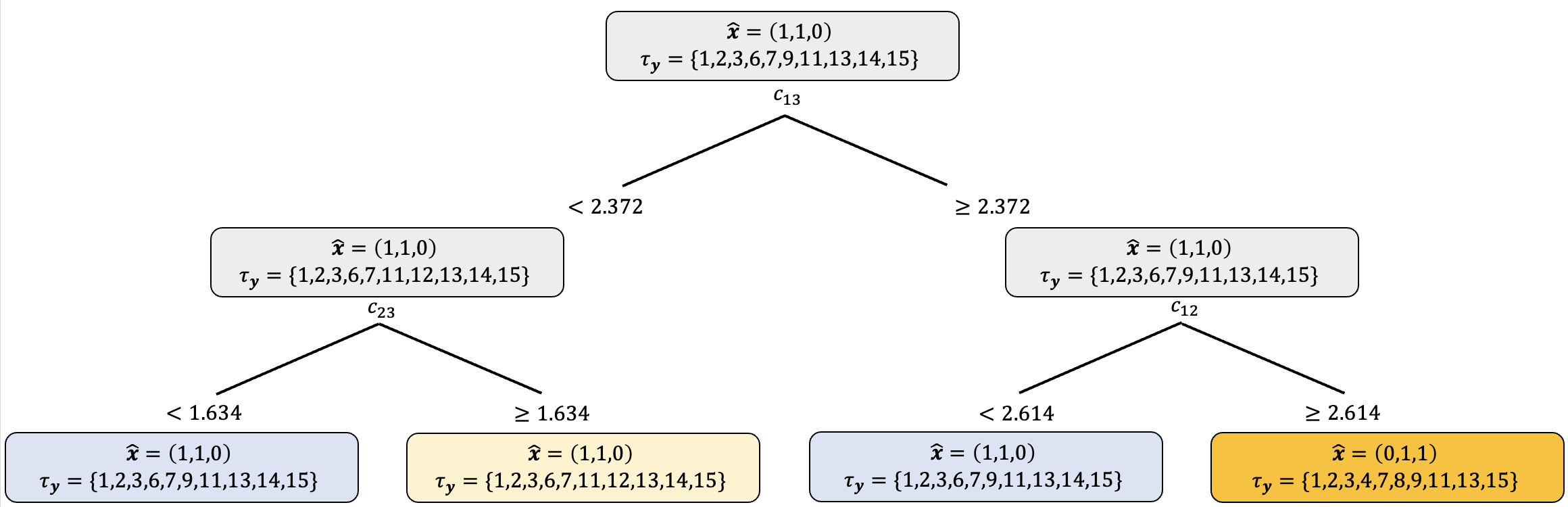}
\caption{Decision tree to predict the optimal strategies for the wait-and-see decisions.}
\label{fig:wait}
\end{figure}

In Figure \ref{fig:here}, we show the decision tree for the here-and-now decisions. Each node contains the predicted here-and-now decision. We can observe how the the cost vector is used to make a construction decision. For example, if $c_{22}$ is smaller than 3.2 and $c_{13}$ is smaller than 2.8, we should build facility both on location 1 and 2 (Note that we always build facility on location 3 regardless of the cost). This makes sense as small value of $c_{22}$ and $c_{13}$ indicates that transporting goods from location 1 and 2 is generally cheap. Likewise, if $c_{22}$ is smaller than 3.2 and $c_{13}$ is larger than 2.8, then we should build facility on location 2 but not on location 1.

% \begin{figure}
%     \centering
%     \includegraphics[scale = 0.3]{image/scenario.png}
%     \caption{Decision tree to predict the optimal strategies for the worst-case scenarios.}
%     \label{fig:scenario}
% \end{figure}

In Figure \ref{fig:scenario}, we show the decision tree for the worst-case scenarios. Each node contains the predicted here-and-now decision and   the associated worst-case scenario. The cost vector is used to make a construction decision, and also predict a worst-case demand that can happen for the construction decision. For example, if $c_{32}$ is larger than 2.662 and $c_{12}$ is larger than 3.411, the worst-case scenario is the scenario in which $d_2$ gets as large as possible within the uncertainty set. A possible interpretation is that large value of $c_{32}$ and $c_{12}$ indicates it is costly to transport goods to destination 2. 
% If $c_{32}$ is larger than 2.662 and $c_{12}$ is smaller than 3.411, the worst-case scenario is the same but the optimal construction decision is to build facility on location 1 instead of location 3.

% \begin{figure}
%     \centering
%     \includegraphics[scale = 0.3]{image/wait.png}
%     \caption{Decision tree to predict the optimal strategies for the wait-and-see decisions.}
%     \label{fig:wait}
% \end{figure}

In Figure \ref{fig:wait}, we show the decision tree for the wait-and-see decisions. Each node contains the predicted here-and-now decision and the indices of the tight constraints. The demand satisfaction constraints are always tight, as we need to minimize cost while satisfying the demand. If $c_{13}$ is larger than 2.372 and $c_{12}$ is larger than 2.614, we should not build facility on location 1. This might be because transporting goods from location 1 is too expensive. Then, constraint 4 is tight, as the right-hand-side of the capacity constraint on location 1 is zero. Constraints 7,8,9 are also tight, as $y_{11} = y_{12} = y_{13} = 0$.  If $c_{13}$ is larger than 2.372 and $c_{12}$ is smaller than 2.614, then we should build facility on location 1 but not on location 3. Then, constraint 6 is tight, as the right-hand-side of the capacity constraint on location 3 is zero. Likewise, constraints 13,14,15 are tight, as $y_{31} = y_{32} = y_{33} = 0$.

{\color{black}
\subsection{Machine Learning Model for Varying Dimensions}
\label{subsec:variability}

In the approach presented earlier, each model is trained for instances with a fixed number of variables and constraints. Now, we introduce a generalized approach so that the trained models can be applied to problems with varying dimensions.

In practice, decision-makers often anticipate encountering a number of contingencies. A contingency refers to a situation where specific decision variables or constraints become irrelevant at the time of here-and-now decision-making. To accommodate this setting, we formally define contingency as a set of decision variables and constraints to be excluded. 

Consider an ARO problem with $n_1$ here-and-now variables, $n_2$ wait-and-see variables, and $m$ constraints for its deterministic version. Then, a contingency can be represented as a triplet $(\mathcal{C}_1, \mathcal{C}_2, \mathcal{C}_3)$, where $\mathcal{C}_1, \mathcal{C}_2, \mathcal{C}_3$ are subsets of $[n_1], [n_2],$ and $[m]$, respectively. They contain the indices of the here-and-now variables, the wait-and-see variables and the constraints to be excluded. An ARO instance can now be associated with both a key parameter $\bm{\theta}$ and a contingency. Solving an ARO instance involves fixing the key parameter to $\bm{\theta}$ and excluding the here-and-now variables, the wait-and-see variables and the constraints whose indices are in $\mathcal{C}_1$, $\mathcal{C}_2$, and $\mathcal{C}_3$, respectively. Then, we apply Algorithm \ref{alg1}.

To integrate this generalization into our framework, we assume that decision-makers have a predefined list of contingencies they expect to encounter, which we refer to as the contingency list. This list serves as an additional input for Algorithm \ref{alg:overview} and \ref{alg:prescriptive}. Given a contingency list, several adjustments are required for these algorithms. In Phase 1, for each contingency in the contingency list, we vary the key parameter to generate multiple instances. This results in training data with diverse combinations of contingencies and key parameters. In Phase 2, the contingencies are integrated into the feature matrix as categorical features. This means that the type of contingency is included as part of the input features for the machine learning models along with the key parameters $\bm{\theta}$. In Phase 3, given a new parameter and a contingency, we remove the variables and constraints specified in the contingency, and then apply the predictions of the machine learning models.

In Phase 2, encoding contingencies as categorical features can be done in various ways. For instance, suppose one must consider all possible combinations of whether to remove or retain $\ell$ different here-and-now variables ${x}_i, i \in [\ell]$. In this case, the contingency list can be represented as $\big\{(\mathcal{C}_1,\emptyset,\emptyset)\big\}_{\mathcal{C}_1 \in 2^{[\ell]}}$. The simplest way to encode contingencies as categorical features is to introduce a single categorical feature representing the type of contingency. Each unique $(\mathcal{C}_1,\emptyset,\emptyset)$ in the contingency list would then correspond to a distinct categorical value. However, this method results in a categorical feature with $2^{\ell}$ distinct values, reflecting the $2^{\ell}$ different contingencies in the contingency list. In such cases, it may be more practical to introduce $\ell$ categorical features. Here, each feature $i \in \ell$ indicates whether $x_i$ is removed or not.

}

{\color{black}

\section{Accelerating Training Data Generation} \label{sec:speedup}

In this section, we introduce a method to expedite the process of generating training data (Phase 1 in Algorithms \ref{alg:overview} and \ref{alg:prescriptive}). As demonstrated in Section \ref{sec:scale}, generating training data in ARO can be computationally intensive. This computational demand may limit the practical applicability of our approach, particularly when frequent retraining of models under various parameter settings is necessary. We address this challenge by drawing inspiration from the principles of online learning. The notation used in this section follows the descriptions provided in Section \ref{sec:CCG}.

\subsection{Motivation}
\label{sec:speedup_motivation}
Conceptually, our work can be viewed as solving a sequence of ARO instances. From this perspective, Algorithm \ref{alg:overview} and \ref{alg:prescriptive} are divided two distinct phases: pure exploration (Phase 1) and pure exploitation (Phase 3). In Phase 1, each ARO instance is solved independently from scratch. We focus solely on collecting data on ARO instances and their solutions without any learning component. Conversely, in Phase 3, we rely entirely on the learned model. This workflow shares similarities with prior works such as \citep{Voice, OPT}, as well as various other learning-based methods to optimization algorithms \citep{cauligi2021coco, nair2021solving, Balcan_Sandholm_Vitercik_2020, doi:10.1287/ijoc.2016.0723}. Our proposed method enhances this process by introducing a more fine-grained approach. Instead of strict divisions between exploration and exploitation, we update prediction models more frequently, gradually reducing the level of exploration over time.

\subsection{Algorithms}
\label{sec:speedup_alg}
We partition Phase 1 into three subphases. The first subphase is the pure exploration stage dedicated to data collection. Using this (potentially very small-sized) data, we train three intermediate models, denoted as $\mathcal{I}_1, \mathcal{I}_2$ and $\mathcal{I}_3$ that are updated throughout Phase 1. These models are utilized to expedite the solution process of Algorithm \ref{alg1}, and the specific manner in which they are utilized distinguishes subphases two and three. First, we outline how Algorithm \ref{alg1} can be expedited using $\mathcal{I}_1, \mathcal{I}_2$, and $\mathcal{I}_3$, followed by a description of the training process.

The goal of $\mathcal{I}_1$ is to expedite the solution process for Problem \eqref{5}. In each iteration $i$ of Algorithm \ref{alg1}, given a set $\mathcal{E}_i$, Problem \eqref{5} is solved to determine the optimal here-and-now decision $\bm{x}_i$ that is robust against the scenarios in $\mathcal{E}_i$. Using the key parameter vector and the latest scenario $\bm{d}_i$ added to $\mathcal{E}_i$ as inputs, $\mathcal{I}_1$ outputs a probability vector indicating the likelihood of each entry of $\bm{x}_i$ being one. The goal of $\mathcal{I}_2$ ($\mathcal{I}_3$) is to provide the initial point $\bm{x}_0$ ($\bm{d}_0$) for Algorithm \ref{alg1} (\ref{alg2}), respectively. 

In the first subphase, we solve ARO instances independently from scratch. This stage focuses solely on gathering data, using random initial points $\bm{x}_0$ for Algorithm \ref{alg1} and three random initial points $\bm{d}_0$ for Algorithm \ref{alg2}. We then train three intermediate models.

In the second subphase, given an ARO instance with the key parameter $\hat{\bm{\theta}}$, we use the predictions of $\mathcal{I}_2$ and $\mathcal{I}_3$ as initial points for Algorithm \ref{alg1} and \ref{alg2}, respectively. In each iteration of Algorithm \ref{alg1}, $\mathcal{I}_1$ predicts the probability of each entry of the here-and-now decision being one. We then set a warm-starting point for the binary variables where the predicted probability is greater than a threshold $p$ to be one, and smaller than $1 - p$ to be zero. The threshold $p$ is set very close to 1, indicating certainty in the model's predictions. This version of Algorithm \ref{alg1} is denoted as Algorithm \ref{alg1_warm}. The intermediate models and the value of $p$ can be updated multiple times as we gather more training data. 

After the second subphase, with more training data, we anticipate improved accuracy in the intermediate models. In the third subphase, we use the predictions of $\mathcal{I}_1$ to partially fix (distinct from warm-starting) the here-and-now variables in each iteration of Algorithm \ref{alg1}. Specifically, we fix variables where the predicted probability exceeds a threshold $p$ or falls below $1 - p$. By fully fixing certain binary variables, our aim is to further expedite Algorithm \ref{alg1_warm}. $\mathcal{I}_2$ and $\mathcal{I}_3$ are utilized in the same manner as in Algorithm \ref{alg1_warm}. This modified version of Algorithm \ref{alg1} is denoted as Algorithm \ref{alg1_fixed}.

The value of $p$ must be chosen carefully, as it directly influences the trade-off between prediction accuracy and the proportion of binary variables that can be fixed or warm-started. A larger $p$ leads to more accurate predictions, but at the expense of being able to fix or warm-start a smaller portion of the binary variables. To determine the value of $p$, we utilize a validation set consisting of 200 data points. We select the smallest $p$ such that the misclassification rate on the validation set, computed only on the entries with probability outputs greater than $p$ or smaller than $1 - p$, is less than 0.00001. 

Now, we elaborate on the training process for $\mathcal{I}_1, \mathcal{I}_2$, and $\mathcal{I}_3$. Assume Algorithm \ref{alg1}, \ref{alg1_warm} or \ref{alg1_fixed} has been applied to the ARO instance associated with a parameter $\bar{\bm{\theta}}$, and it terminated after $J$ iterations. Then, the training data extracted from this single instance for $\mathcal{I}_1$ is  $\big((\bar{\bm{\theta}}, \bm{d}_j), \bm{x}_j \big)_{j=1}^{J}$. The training data for $\mathcal{I}_2$ is $\big(\bar{\bm{\theta}}, \bm{x}_J \big)$, while for $\mathcal{I}_3$ it is $\big(\bar{\bm{\theta}}, \bm{d}_J \big)$. $\mathcal{I}_1$ and $\mathcal{I}_2$ are binary classifiers that predict whether each entry of the here-and-now decision is one, while $\mathcal{I}_3$ is a multiclass classifier similar to the models in Algorithm \ref{alg:overview}. While the prediction targets of $\mathcal{I}_1$ and $\mathcal{I}_2$ resemble those described in Algorithm \ref{alg:overview} and \ref{alg:prescriptive}, the models in those algorithms predict the entire here-and-now decision vector as a unified bundle. On the contrary, $\mathcal{I}_1$ and $\mathcal{I}_2$ are binary classifiers that predict whether each entry of the here-and-now variable is zero or one individually. Hence, neural networks are specifically chosen due to the high-dimensionality of the prediction target.

\paragraph{Remark}  In Section \ref{sec:accel}, we show that the solution quality of Algorithm \ref{alg1_warm} and \ref{alg1_fixed} remains practically identical to Algorithm \ref{alg1}.  Even if solutions are of poor quality, however, it does not pose a significant challenge to our main approach in Algorithm \ref{alg:prescriptive}. This is because during the computation of the reward matrix, entries associated with poor solutions will be assigned high suboptimality.

\begin{algorithm}
\small
 \KwInput{Problem \eqref{1}, $\mathcal{I}_1$, $\mathcal{I}_2$, $\mathcal{I}_3$,  $\epsilon_1$, $\epsilon_2$, $p$}
\KwOutput{$\tilde{\bm{x}}^*$,$\tilde{\bm{d}}^*$} 
 \textbf{Initialization}: $i = 0$,  $\mathcal{E}_0 = \emptyset$,  UB = $\infty$, LB = $-\infty$\\
  
 \vspace{2mm}
  $\mathcal{I}_2$ predicts $\bm{x}_0$ and $\mathcal{I}_3$ predicts $\hat{\bm{d}}_0$ \\
 \While{UB -LB $\geq \epsilon_1$}{
  \If{$i = 0$}{ Evaluate $\mathcal{Q}(\bm{x}_i)$ using $\hat{\bm{d}}_0$ as the initial point in Algorithm \ref{alg2} to get $\tilde{\mathcal{Q}}(\bm{x}_i)$ and a solution $\bm{d}_i$. \\
  $\mathcal{E}_{i+1} \leftarrow \mathcal{E}_{i} \bigcup \{\bm{d}_i\}$\\
  $i$ $\leftarrow$ $i+1$
  }
  \Else{$\mathcal{I}_1$ outputs a probability vector. \\
  Solve \eqref{5} with the extreme points in $\mathcal{E}_i$. The binary variables whose corresponding entries in the probability vector that are greater than $p$ and smaller than $1 - p $ are warm-started at 1 and 0, respectively. Denote the solutions as $\bm{x}_i$ and $\bm{\alpha}_i$. \\
  LB $\leftarrow$ $\bm{\alpha}_i$\\
  Evaluate $\mathcal{Q}(\bm{x}_i)$ using $\hat{\bm{d}}_0$ as the initial point in Algorithm \ref{alg2} to get $\tilde{\mathcal{Q}}(\bm{x}_i)$ and a solution $\bm{d}_i$.\\
  UB $\leftarrow$ $\tilde{\mathcal{Q}}(\bm{x}_i)$\\
  $\mathcal{E}_{i+1} \leftarrow \mathcal{E}_{i} \bigcup \{\bm{d}_i\}$\\
  $i$ $\leftarrow$ $i+1$}{
  
  }} 
  $\tilde{\bm{x}}^* \leftarrow \bm{x}_i$ \\
  $\tilde{\bm{d}}^* \leftarrow \bm{d}_i$
 \caption{Column and Constraint Generation with Warm Start}
 \label{alg1_warm}
\end{algorithm}

\begin{algorithm}
\small
 \KwInput{Problem \eqref{1},$\mathcal{I}_1$, $\mathcal{I}_2$, $\mathcal{I}_3$, $\epsilon_1$, $\epsilon_2$, $p$}
\KwOutput{$\tilde{\bm{x}}^*$,$\tilde{\bm{d}}^*$} 
 \textbf{Initialization}: $i = 0$, $\bm{x}_0$, $\mathcal{E}_0 = \emptyset$,  UB = $\infty$, LB = $-\infty$ \\
\vspace{2mm}
  $\mathcal{I}_2$ predicts $\bm{x}_0$ and $\mathcal{I}_3$ predicts $\hat{\bm{d}}_0$ \\
 \While{UB -LB $\geq \epsilon_1$}{
  \If{$i = 0$}{ Evaluate $\mathcal{Q}(\bm{x}_i)$ using $\hat{\bm{d}}_0$ as the initial point in Algorithm \ref{alg2} to get $\tilde{\mathcal{Q}}(\bm{x}_i)$ and a solution $\bm{d}_i$. \\
  $\mathcal{E}_{i+1} \leftarrow \mathcal{E}_{i} \bigcup \{\bm{d}_i\}$\\
  $i$ $\leftarrow$ $i+1$
  }
  \Else{$\mathcal{I}_1$ outputs a probability vector. \\
  Solve \eqref{5} with the extreme points in $\mathcal{E}_i$. The binary variables whose corresponding entries in the probability vector that are greater than $p$ and smaller than $1 - p$ are fixed at 1 and 0, respectively. Denote the solutions as $\bm{x}_i$ and $\bm{\alpha}_i$. \\
  LB $\leftarrow$ $\bm{\alpha}_i$\\
  Evaluate $\mathcal{Q}(\bm{x}_i)$ using $\hat{\bm{d}}_0$ as the initial point in Algorithm \ref{alg2} to get $\tilde{\mathcal{Q}}(\bm{x}_i)$ and a solution $\bm{d}_i$.\\
  UB $\leftarrow$ $\tilde{\mathcal{Q}}(\bm{x}_i)$\\
  $\mathcal{E}_{i+1} \leftarrow \mathcal{E}_{i} \bigcup \{\bm{d}_i\}$\\
  $i$ $\leftarrow$ $i+1$}{
  
  }} 
  $\tilde{\bm{x}}^* \leftarrow \bm{x}_i$ \\
  $\tilde{\bm{d}}^* \leftarrow \bm{d}_i$
 \caption{Column and Constraint Generation with Warm Start and Fixed Variables}
 \label{alg1_fixed}
\end{algorithm}

}

\section{Partitioning Algorithm} \label{sec5}
In this section, we propose an algorithm to reduce the number of distinct strategies for the wait-and-see decisions in the training set. In the computational experiments, we have observed that as the size of the uncertainty sets gets large, the number of distinct strategies for the wait-and-see variables in the training set can get prohibitively large. While a similar issue is discussed in \citep{Prune}, there is a subtle difference in our context. For MICO problems, there is no notion of uncertainty set. Therefore, the number of distinct strategies is controlled by the support of the training distribution. If the number becomes too large for a distribution of interest, one can partition its support into multiple smaller regions and train a machine learning model for each region. However, this approach does not directly translate to ARO, as the number of distinct strategies depends on both the training distribution and the size of the uncertainty set. We cannot simply reduce or partition the uncertainty set, because this leads to less robust solutions. Moreover, the pruning algorithm described in \citep[Section~4.3]{Prune} is often insufficient to handle the large number of strategies encountered in our numerical experiments with large uncertainty sets. However, the algorithm we develop in this section can reduce the number effectively. 

The high-level idea is that instead of trying to identify the tight constraints, we try to identify a subset of the redundant constraints. We can optimize excluding these constraints and still get the optimal solution to the original problem. 

Before giving a formal description of the algorithm, we first provide a small motivating example. Consider the following hypothetical setting. We are given a deterministic continuous optimization problem with four constraints, each denoted as constraint 1,2,3,4, respectively. In Phase 1, we generate four training parameters, $\bm{\theta}_i, i \in [4]$, and solve the associated instances to optimality. The tight constraints (which also defines the optimal strategy as the problem of interest is continuous) for each instance is $\tau(\bm{\theta}_i) = \{i\}$. That is, the optimal strategies of the training instances are all different, resulting in four distinct target classes. Learning in this setting is challenging, as the number of distinct target classes is equal to the number of training instances. Using $\bm{\tau}$ to denote the set of tight constraints found in the training set, we get $\bm{\tau} = \Big\{ \{1\},\{2\},\{3\},\{4 \}  \Big\}$. 

In the algorithm we propose, we first need to define a partition of the set $\bm{\tau}$. In this example, we define the partition as $\mathcal{P} = \bigg\{ \Big\{\{1\},\{2\}\Big\}, \Big\{\{3\},\{4\}\Big\}  \bigg\}$. For each cell in $\mathcal{P}$, we compute the union of its elements. The unions are $\{1,2\}$ and $\{3,4\}$ under the partition we defined. Then, for the parameters $\bm{\theta}_1$ and $\bm{\theta}_2$, we redefine their prediction targets as  $\{1,2\}$. We can still compute the optimal solutions of the the parameters $\bm{\theta}_1$ and $\bm{\theta}_2$ by imposing the constraints $\{1,2\}$ only. Likewise, for $\bm{\theta}_3$ and $\bm{\theta}_4$, we redefine their prediction targets as $\{3,4\}$. Once we redefine the prediction targets following this procedure, the number of distinct prediction targets is reduced to two. A downside might be that given a new instance, prediction of a trained model now contains two constraints instead of one. This can undermine the computational efficiency we could have gained by imposing just a single constraint. The partitioning algorithm we propose is a generalization of this procedure to two-stage ARO with binary here-and-now variables.

Assume we have $N$ ARO instances and the corresponding optimal strategies $(s_1, \dots, s_N)$, where $s_i = (\bm{x}^*_i, \tau_{\bm{y},i}), i \in [N]$. Let $\bm{\tau} = \{\tau_{\bm{y},1}, \tau_{\bm{y},2}, \dots, \tau_{\bm{y},N}\}$ be the set of tight constraints in our training set. Without loss of generality, we assume $\bm{\tau} = \{\tau_{\bm{y},1}, \tau_{\bm{y},2}, \dots, \tau_{\bm{y},M}\}$, where $\tau_{\bm{y},i} \neq \tau_{\bm{y},j}$ if $i \neq j$, and $\bm{\tau}$ is sorted in the order such that if $i < j$, then $\tau_{\bm{y},i}$ occurred more frequently than $\tau_{\bm{y},j}$ in the training set (ties are broken arbitrarily). Note that $M \leq N$, since the optimal strategies might overlap. We divide $\bm{\tau}$ into K partitions $ \mathcal{P} = \{\mathcal{P}_1, \mathcal{P}_2, \dots, \mathcal{P}_K\} \; (K \leq M)$, and compute the union of the elements in each cell. We use $u_i$ to denote the union of the elements in the cell that $\tau_{\bm{y},i}$ originally belonged to. For the $i_{th}$ instance, we replace its prediction target with $(\bm{x}^*_i, u_i)$. Algorithm \ref{alg_partition} provides a formal description. 
 
In our implementation in Section \ref{sec:numerical experiment}, we define the partition of $\bm{\tau}$ the following way. We let $\mathcal{P}_i = \{\tau_{\bm{y},i} \}, i \in [K-1]$, and $\mathcal{P}_K = \{\tau_{\bm{y},K}, \dots, \tau_{\bm{y},M}\}$. In other words, we let $K-1$ most
frequently occurring tight constraints form their own partitions with a single element. We combine the rest of the tight constraints to form the $K_{th}$ cell, resulting in $K$ cells in total. We denote $\bar{u} = \bigcup_{\tau_{\bm{y},j} \in \mathcal{P}_K}\tau_{\bm{y},j} $ as the union constraints.

As mentioned above, using the union of the tight constraints can undermine the computational efficiency that we could have gained by imposing only the exact tight constraints. Another concern might be that as we are artificially redefining the prediction targets in the data set, training an accurate prediction model might become challenging. However, empirically, tight constraints of different instances mostly overlap. Hence, the cardinality of the union constraints is in general similar to the cardinality of the individual tight constraints. Furthermore, even after applying Algorithm \ref{alg_partition} to the data set, accurate models can be trained. We demonstrate these points in Section \ref{sec:numerical experiment}.

\begin{algorithm}
\small
\SetAlgoLined
\KwOut{$\bigg\{ \Big(\bm{\theta_i}, (\bm{x}^*_i,u_i) \Big) \bigg\}_
{i=1}^N$}
 \bf{Input}: $\bigg\{ \Big(\bm{\theta_i}, (\bm{x}^*_i,\tau_{\bm{y},i}) \Big) \bigg\}_{i=1}^N$, $\mathcal{P} = \{\mathcal{P}_1, \mathcal{P}_2, \dots, \mathcal{P}_K\}$ \\
  \For{$\mathcal{P}_i \in \mathcal{P}$}{
  $\bar{\tau}_{\bm{y},i}  \leftarrow \bigcup_{\tau_{\bm{y},j} \in \mathcal{P}_i} \tau_{\bm{y},j}$
  }

 \For{$i=1$ \KwTo $N$}{
 \For{$j=1$ \KwTo $K$}{
 \If{$\tau_{\bm{y},i} \in \mathcal{P}_j $}{
 $u_i \leftarrow \bar{\tau}_{\bm{y},j}$ \\
 break 
 }
 }
 }
 \caption{Partitioning Algorithm}
 \label{alg_partition}
\end{algorithm}

\section{Computational Experiments}
\label{sec:numerical experiment}

In this section, we provide the results of the computational experiments on synthetic and real-world problems. We evaluate the quality of the predicted strategies and also analyze the relative speed-up of our approach compared with Algorithm \ref{alg1}. {\color{black} We also demonstrate the effectiveness of Algorithm \ref{alg1_warm}, \ref{alg1_fixed} and \ref{alg_partition}. In the Supplementary Material, we offer further insights into the performance of our approach across a range of scenarios. This includes a report on the offline computation time of our method, as well as an analysis of its performance under varying sizes of uncertainty sets, training data, and distributional shifts.} Identical to the experiment in Section \ref{sec:scale}, the experiment in this section was executed in Julia 1.4.1 on a MacBook Pro with 2.6 GHz Intel Core i7 CPU and 16GB of RAM. Likewise, all deterministic optimization problems involved are solved with Gurobi. The software for OPT is available from \citet{InterpretableAI}.

\subsection{Problem Description}
\label{sec:description}

We describe the synthetic and real-world problems that we test our approach on. We also provide sample generation details and the uncertainty sets used.

\paragraph{Facility Location}

We consider the facility location problem introduced in Section \ref{subsec:facility}. We use the polyhedral uncertainty set defined as  $\mathcal{D} = \{\bm{d} \mid \sum_{i}d_i \leq \Gamma, 4 \leq d_i \leq 6 \}$. The feature vector is $\bm{f}$.  For the case with $n =7$, we sample $\bm{f}$ from the ball $B(\bar{\bm{f}}, 3)$, where $\bar{f_i}$ is sampled from $U(2,12)$ and fixed. For all other cases, we sample $\bm{f}$ from $B(\bar{\bm{f}}, 1.5)$, where $\bar{f_i}$ is sampled from $U(2,22)$ and fixed.  We sample $p_i$ from $U(8,18)$ and $c_i$ from $U(2,4)$.

\paragraph{Inventory Control}

Consider the multi-item inventory control problem, where ordering decisions can be partially made after the demand is realized. There are $n$ different items to order, with three different ways to order each item. For each item $i, i \in [n]$, we can order a fixed lot size of $l_i$ at the unit cost of $c_i^1$ or order a fixed lot size of $l_i$ at the unit cost of $c_i^2$ before the demand is realized. After we see the demand, we can order any amount $y_i$ at the unit cost of $c_i^3$. We must also pay storage and disposal cost of $c^4$ for the remaining stock after the demand is satisfied. We set $c_i^1,c_i^2 \leq c_i^3 \leq c_i^4$ to avoid trivial solutions. The here-and-now binary variables to decide whether we order fixed lot sizes with the cost $c_i^1$ and $c_i^2$ before seeing the demand are denoted $x_i^1$ and $x_i^2$, respectively. The wait-and-see variable is $y_i$. The exact formulation is as follows.

\begin{alignat*}{3}
&\underset{\bm{y}(\cdot),\bm{x}^1, \bm{x}^2}\min\quad && \underset{\bm{d} \in \mathcal{D}}\max \quad \sum_{i = 1}^{n} c_i^{1}l_{i}x_i^{1} + \sum_{i = 1}^{n} c_i^{2}l_{i}x_i^{2} + && \sum_{i = 1}^{n} c_i^{3}y_i(\bm{d})  +   c^4\sum_{i = 1}^{n}[l_{i}x_{i}^1 + l_{i}x_{i}^2+y_i (\bm{d})- d_i ] \\
& \text{s.t.} && l_{i}x_{i}^1 + l_{i}x_i^2 + y_i(\bm{d}) - d_i \geq 0, && \quad \forall {\bm{d}\in\mathcal{D}}, \quad \forall {i\in [n]},  \nonumber \\
& && y_{i}(\bm{d}) \geq 0, && \quad  \forall {\bm{d}\in\mathcal{D}}, \quad \forall {i\in [n]}, \nonumber \\
& && \bm{x}^1,\bm{x}^2 \in \{0,1\}^{n}. \nonumber
\end{alignat*}

We use the uncertainty set defined as $\mathcal{D} = \{\bm{d} \mid \norm{\bm{d} - 50}_{1} \leq \Gamma \}$. For the problem with $n=25$, feature vectors are $\bm{c}^2$ and $\bm{c}^3$. We sample $\bm{c}^2$ from $B(\bar{\bm{c}^2}, 5)$, where $\bar{c_i^2}$ is sampled from $U(40,60)$ and fixed. We sample $\bm{c}^3$ from $B(\bar{\bm{c}^3}, 5)$, where $\bar{c_i^3}$ is sampled from $U(60,80)$ and fixed. For the larger problems, the feature vector is $\bm{c}^3$. We sample $\bm{c}^3$ from $B(\bar{\bm{c}^3}, 2)$, where $\bar{c_i^3}$ is sampled from $U(60,80)$ and fixed. We sample $l_i$ from $U(20,30)$, $c_i^1$ from $U(40,60)$ and fix $c_4 = 60$.

\paragraph{Unit Commitment}

We consider the unit commitment problem described in Section \ref{sec:scale}. We give the complete formulation of the deterministic version in the Supplementary Material, and the data is taken from \citep{UCdata}. This problem is analogous to the facility location problem, but with more complicated constraints. We use the budget uncertainty set defined as $\mathcal{D} = \{\bm{d} \mid \sum_{i}\abs{\frac{d_i - \bar{d_i}}{0.1 \times \bar{d_i}}} \leq \Gamma, \abs{d_i - \bar{d_i}} \leq 0.1 \times \bar{d_i}  \}$, where $\bar{\bm{d}}$ is the original data. The feature vector is the coefficient vector $\bm{b}$ of the production cost function. The parameters are sampled from the ball with radius 1.5, and the center of the ball is the original data.

\subsection{Experimental Design}
\label{subsec:experimental_setup}

We conduct two sets of experiments. In the first set of experiments, we generate and solve ARO instances to near-optimality in Phase 1 using tight tolerances for Algorithm \ref{alg1} and \ref{alg2}. Then, we use XGBOOST \citep{XGBoost} for the classification approach in Algorithm \ref{alg:overview}, and compare its performance with OPT (Algorithm \ref{alg:prescriptive}). There are two main reasons behind this experimental design. First, we aim to demonstrate the effectiveness of our approach regardless of the machine learning method used. Second, we aim to evaluate the trade-off between interpretability and prediction accuracy. XGBOOST is known for its high performance on various prediction tasks but lacks interpretability compared to OPT. In contrast, OPT is highly interpretable due to its simple decision tree structure but may have weaker prediction accuracy compared to XGBOOST. By comparing these two methods, we aim to analyze the cost we have to pay to gain interpretability. {\color{black} We also analyze the effectiveness of Algorithm \ref{alg_partition} and the generalization described in Section \ref{subsec:variability}.} We provide the results in Section \ref{sec:numerical_acc},\ref{sec:numerical_partition} and \ref{sec:numerical_variability}.

In the second set of experiments, we generate suboptimal strategies in Phase 1 to solve large scale unit commitment problems. As shown in Section \ref{sec:scale}, solving such problems can be computationally challenging. As a result, generating a training set in Phase 1 can be a significant computational burden. To overcome this issue, we use more relaxed tolerances for Algorithm \ref{alg1} and \ref{alg2} to terminate earlier. In Section \ref{sec:largescale}, we demonstrate that Algorithm \ref{alg:prescriptive} can still find high quality solutions for large scale unit commitment problems. {\color{black} In Section \ref{sec:accel}, we demonstrate the effectiveness of Algorithm \ref{alg1_warm} and \ref{alg1_fixed} to further expedite training set generation.}

Furthermore, \citet{Voice, Prune} propose using multiple predictions of the trained model in Phase 3. Classification algorithms generate a likelihood vector where each entry represents the likelihood of a particular label being the true label for a data point. Hence, multiple most promising predictions can be identified using this vector. Similarly, OPT can output multiple best strategies \citep{OPT}. We can evaluate all of these predictions in parallel by computing their infeasibilities and suboptimalities to choose the best one. The drawback of this approach is that the evaluation process requires additional computation in Phase 3. For both experiments, we use multiple predictions of OPT, only if using just a single prediction does not result in perfect accuracy. In this case, we provide separate tables to analyze the performance improvement and the additional computational burden. We use $k$ to denote the number of predictions we use in Phase 3.

\subsection{Solving ARO with Near-Optimal Strategies}
\label{sec:numerical_acc}

In this section, we generate and solve ARO instances to near-optimality in Phase 1. Throughout the entire experiment, we set $\epsilon_1 = 0.001$ and $\epsilon_2 = 0.001$ for Algorithm \ref{alg1} and \ref{alg2}, respectively. The optimality gap of Gurobi is fixed at its default value of 0.0001. When using OPT, we use the entire set of strategies found to ensure a fair comparison with XGBoost. In other words, we set  $Q_1 = |\mathcal{S}_{\bm{x}}| , Q_2 = |\mathcal{S}_{\bm{d}}|$ and $ Q_3 = |\mathcal{S}_{\bm{y}}|$. For both XGBoost and OPT, we minimize the hyperparameter tuning process and grid search over the maximum depths 5 and 10. 

Table \ref{table:facility}, \ref{table:inventory}, \ref{table:uc} contain the experiment results on the facility location, the inventory control and the unit commitment problem, respectively. For the experiments reported in these tables, we only use a single prediction in Phase 3 ($k = 1$). Table \ref{table:facility_multiple}, \ref{table:inventory_multiple}, \ref{table:UC_multiple} contain the experiment results using multiple predictions of OPT in Phase 3 ($k \geq 1$). As mentioned above, we experiment with $k \geq 1$ only if the prediction accuracy with $k=1$ is not perfect. We report how the performance changes as we increase $k$. 

\paragraph{Table Notations}

Total $N$ ARO instances are generated, which are randomly split into the training set (70\%) and the test set (30\%). Columns $n$ and $m$ contain the parameters that define the problem size and column $\Gamma$ contains the parameter that determines the size of the uncertainty sets. In the Accuracy column, we report the percentage of accurate predictions on the test set, rounded up to the second decimal place. For all three prediction targets, we consider a prediction accurate if it is feasible and the suboptimality is smaller than 0.0001. In the Infeasibility column, we report the percentage of infeasible predictions on the test set. We report the maximum suboptimality among the feasible predictions in the column $sub_{max}$. In the $\abs{\mathcal{S}}$ column, we report the number of distinct strategies found in the training set. In case we used Algorithm \ref{alg_partition} to reduce this number, we report the number we get by applying Algorithm \ref{alg_partition}, not the original number. We provide the original number of strategies and further analysis of Algorithm \ref{alg_partition} in Section \ref{sec:numerical_partition}. In the $t_{ratio}$ column, we report the computation time it takes to obtain the solution from scratch using Algorithm \ref{alg1}, divided by the computation time of our approach. It is rounded up to the nearest integer. 

\begin{table*}\centering
\small
\footnotesize

\begin{tabular}{ccccccccccc}
\toprule
Target  &   $n$   & $m$  &$\Gamma$ & Learner & Accuracy  & Infeasibility & $sub_{max}$ &  $\abs{\mathcal{S}}$  &  $N$ & $t_{ratio}$    \\ 
\midrule
$s_{\bm{x}}$  &  \multirow{3}{*}{7} & \multirow{3}{*}{7}   & \multirow{3}{*}{38}  &   \multirow{3}{*}{OPT} & 1.00 & 0   & 0.0070  &2 &  20000 & 1666\\ 
$s_{\bm{d}}$ &  &  &   &   & 1.00 & 0   & 0.0000       & 4   &  20000& 1538 \\ 
$s_{\bm{y}}$ &  & &   &   &  0.93 & 0   & 0.0070       & 23   & 20000& 34\\
\hline
$s_{\bm{x}}$ &  \multirow{3}{*}{7} & \multirow{3}{*}{7}   & \multirow{3}{*}{38}  &   \multirow{3}{*}{XGB} & 1.00 & 0   & 0.0000  &2 &  20000 & 34\\ 
$s_{\bm{d}}$ &  &  &   &   & 1.00 & 0   & 0.0000       & 4   &  20000& 32\\ 
$s_{\bm{y}}$ &  &  &   &   &  0.97 & 0   & 0.0010       & 23   & 20000& 17\\

\hline

$s_{\bm{x}}$ &  \multirow{3}{*}{80} & \multirow{3}{*}{60}   &  \multirow{3}{*}{241}  & \multirow{3}{*}{OPT} & 0.99 &  0 & 0.0004 &    10 & 20000 & 33333\\ 
$s_{\bm{d}}$ &   &  &    &  & 0.99 &  0 &  0.0004    &     10  & 20000 &  36363\\ 
$s_{\bm{y}}$ &   &  &   &   &  0.98 & 0   & 0.0004        & 22    & 20000& 21\\
\hline 
$s_{\bm{x}}$ &   \multirow{3}{*}{80} & \multirow{3}{*}{60}   & \multirow{3}{*}{241}  &   \multirow{3}{*}{XGB} & 1.00 & 0   & 0.0000  & 10 &  20000 & 276\\ 
$s_{\bm{d}}$ &   &  &   &   & 0.99 & 0   & 0.0004       & 10   &  20000& 278\\ 
$s_{\bm{y}}$ &   &  &   &   &  0.95 & 0   & 0.0004       & 22   & 20000& 21\\

\hline
$s_{\bm{x}}$ &  \multirow{3}{*}{200} & \multirow{3}{*}{150}   &  \multirow{3}{*}{601}  & \multirow{3}{*}{OPT} & 1.00 & 0  & 0.0002 &    10 & 25000 & $3.75 \times 10^5$\\ 
$s_{\bm{d}}$ & &  &   &   & 0.99 & 0   &  0.0002    &     10  & 25000 &  $4.06 \times 10^5$\\ 
$s_{\bm{y}}$ &  &  &  &    &  0.99 &  0  & 0.0002        & 10    & 25000& 37\\
\hline 
$s_{\bm{x}}$ &   \multirow{3}{*}{200} & \multirow{3}{*}{150}   & \multirow{3}{*}{601}  &   \multirow{3}{*}{XGB} & 1.00 & 0   & 0.0002  & 10 &  25000 & 1036\\ 
$s_{\bm{d}}$ & &  &   &   & 0.98 & 0   & 0.0002       & 10   &  25000& 880\\ 
$s_{\bm{y}}$ & &  &   &   &  0.97 & 0   & 0.0002       & 10   & 25000& 35 \\

\hline
$s_{\bm{x}}$ &   \multirow{3}{*}{200} & \multirow{3}{*}{150}   &  \multirow{3}{*}{751}  & \multirow{3}{*}{OPT} & 1.00 &0 &0.0000  &    42  & 25000 & $3.09 \times 10^6$\\ 
$s_{\bm{d}}$ &  &  &   &   & 1.00 &  0  &  0.0000    &     42  & 25000 & $3.37 \times 10^6$\\ 
$s_{\bm{y}}$ &  &  &   &   &  1.00 &  0  & 0.0000        &   42  & 25000& 186\\
\hline
$s_{\bm{x}}$ &  \multirow{3}{*}{200} & \multirow{3}{*}{150}   & \multirow{3}{*}{751}  &   \multirow{3}{*}{XGB} & 0.99 & 0   & 0.0020  & 42 &  25000 & 9568\\ 
$s_{\bm{d}}$ & &  &   &   & 0.99 & 0   & 0.0020       & 42  &  25000& 12783\\ 
$s_{\bm{y}}$ & &  &   &   &  0.99 & 0   & 0.0020       & 42   & 25000& 183\\

\bottomrule
\end{tabular}
\caption{Numerical results for the facility location problem with $k = 1$.}
\label{table:facility}
\end{table*}

\begin{table*}\centering
\small
\footnotesize

\begin{tabular}{cccccccccc}
\toprule
  Target&   n    &$\Gamma$ & Learner & Accuracy  & Infeasibility & $sub_{max}$ &  $\abs{\mathcal{S}}$  &  $N$ & $t_{ratio}$    \\ 
\midrule
$s_{\bm{x}}$ & \multirow{3}{*}{25} & \multirow{3}{*}{10}  &   \multirow{3}{*}{OPT} & 1.00 & 0   & 0.0004  & 12 &  40000 & 1220 \\ 
$s_{\bm{d}}$ &   &    &   & 1.00 & 0   & 0.0004       & 28   &  40000& 1126 \\ 
$s_{\bm{y}}$  &   &     &   &  0.99 & 0   & 0.0004       & 30   & 40000& 38\\
\hline
$s_{\bm{x}}$ &   \multirow{3}{*}{25} & \multirow{3}{*}{10}  &   \multirow{3}{*}{XGB} & 1.00 & 0   & 0.0000  &12 &  40000 & 13\\ 
$s_{\bm{d}}$ &   &  &   & 1.00 & 0   & 0.0000       & 28   &  40000& 14\\ 
$s_{\bm{y}}$ &   &     &   &  0.99 & 0  & 0.0004       & 30   & 40000& 8\\

\hline

$s_{\bm{x}}$ &   \multirow{3}{*}{600} &  \multirow{3}{*}{10}  & \multirow{3}{*}{OPT} & 1.00 &  0 & 0.0000 &    17 & 60000 & 17241\\ 
$s_{\bm{d}}$ &   &      &  & 1.00 &  0 &  0.0000    &     30  & 60000 & 14589\\ 
$s_{\bm{y}}$  &   &  &      &  1.00 & 0   & 0.0000        & 48     & 60000& 78\\
\hline 
$s_{\bm{x}}$ &   \multirow{3}{*}{600}  & \multirow{3}{*}{10}  &   \multirow{3}{*}{XGB} & 1.00 & 0   & 0.0000  & 17 &  60000 & 21\\ 
$s_{\bm{d}}$ &  &   &   & 1.00 & 0   & 0.0000       & 30   &  60000& 22\\ 
$s_{\bm{y}}$  &   &     &   &  1.00 & 0   & 0.0000       & 48   & 60000& 14\\

\hline
$s_{\bm{x}}$ & \multirow{3}{*}{1000}   &  \multirow{3}{*}{10}  & \multirow{3}{*}{OPT} & 1.00 & 0  & 0.0000 &    9 & 60000 & 12838 \\ 
$s_{\bm{d}}$ &  &     &   & 1.00 & 0   &  0.0000    &     27  & 60000 &  11851\\ 
$s_{\bm{y}}$  &  &   &    &  1.00 &  0  & 0.0000        & 11    & 60000& 84\\
\hline 
$s_{\bm{x}}$ &  \multirow{3}{*}{1000}   & \multirow{3}{*}{10}  &   \multirow{3}{*}{XGB} & 1.00 & 0   & 0.0000  & 9 &  60000 & 13\\ 
$s_{\bm{d}}$ &   &     &   & 0.99 & 0   & 0.0000       & 27   &  60000& 13\\ 
$s_{\bm{y}}$  &     &   &   &  1.00 & 0   & 0.0000       & 11   & 60000& 13 \\

\hline
$s_{\bm{x}}$ &  \multirow{3}{*}{1000}   &  \multirow{3}{*}{45}  & \multirow{3}{*}{OPT} & 1.00 &0 &0.0000  &   7  & 60000 & 25480\\ 
$s_{\bm{d}}$ &   &  &      & 1.00 &  0  &  0.0000    &     30  & 60000 & 23520\\ 
$s_{\bm{y}}$  &  &  &      &  1.00 &  0  & 0.0000        &   7  & 60000& 94\\
\hline
$s_{\bm{x}}$ & \multirow{3}{*}{1000}   & \multirow{3}{*}{45}  &   \multirow{3}{*}{XGB} & 1.00 & 0   & 0.0000  & 7 &  60000 & 21\\ 
$s_{\bm{d}}$ &   &  &      & 1.00 & 0  & 0.0000       & 30  &  60000 & 24\\ 
$s_{\bm{y}}$  &   &     &   &  1.00 & 0   & 0.0000       &7   & 25000& 17\\

\bottomrule
\end{tabular}
\caption{Numerical results for the inventory control problem with $k=1$.}
\label{table:inventory}
\end{table*}

\begin{table*}\centering
\small
\footnotesize

\begin{tabular}{ccccccccccc}
\toprule
 Target &   $n$   & $m$ &$\Gamma$ & Learner & Accuracy  & Infeasibility & $sub_{max}$ &  $\abs{\mathcal{S}}$  &  $N$ & $t_{ratio}$    \\ 
\midrule
$s_{\bm{x}}$ &   \multirow{3}{*}{10} & \multirow{3}{*}{24}   & \multirow{3}{*}{0.1}  &   \multirow{3}{*}{OPT} & 0.97 & 0   & 0.0010  &17 &  20000 & 88137\\ 
$s_{\bm{d}}$ &   &  &   &   & 0.97 & 0   & 0.0010      & 24   &  20000& $1.30 \times 10^5$ \\ 
$s_{\bm{y}}$ &   &  &   &   &  0.96 & 0   & 0.0010       & 32   & 20000& 119\\
\hline
$s_{\bm{x}}$ &  \multirow{3}{*}{10} & \multirow{3}{*}{24}   & \multirow{3}{*}{0.1}  &   \multirow{3}{*}{XGB} & 1.00 & 0   & 0.0000  &17 &  20000 & 5137\\ 
$s_{\bm{d}}$ &   &  &   &   & 0.98 & 0   & 0.0040       & 24   &  20000&  6445\\ 
$s_{\bm{y}}$ &   &  &   &   &  0.93 & 0   & 0.0040       & 32   & 20000& 277\\

\hline

$s_{\bm{x}}$ &  \multirow{3}{*}{10} & \multirow{3}{*}{24}   & \multirow{3}{*}{2}  &   \multirow{3}{*}{OPT} & 1.00 & 0   & 0.0000  &9 &  15000 & 93318\\ 
$s_{\bm{d}}$ &   &  &   &   & 1.00 & 0   & 0.0000       & 9   &  15000&  87228 \\ 
$s_{\bm{y}}$ &  &  &   &   &  1.00 & 0   & 0.0000       & 9   & 15000& 296\\
\hline
$s_{\bm{x}}$ & \multirow{3}{*}{10} & \multirow{3}{*}{24}   & \multirow{3}{*}{2}  &   \multirow{3}{*}{XGB} & 1.00 & 0   & 0.0004  &9 &  15000 & 5300\\ 
$s_{\bm{d}}$ &  &  &   &   & 1.00 & 0   & 0.0000       & 9   &  15000&  6389\\ 
$s_{\bm{y}}$ &  &  &   &   &  1.00 & 0   & 0.0000       & 9   & 15000& 293\\

\bottomrule
\end{tabular}
\caption{Numerical results for the unit commitment problem with $k=1$.}
\label{table:uc}
\end{table*}

\begin{table*}\centering
\footnotesize
\begin{tabular}{cccccccccccc}
\toprule
 Target & $k$ &  n   & m  &$\Gamma$  & Accuracy  & Infeasibility & $sub_{max}$ &  $\abs{\mathcal{S}}$  &  $N$ & $t_{ratio}$    \\ 
  \midrule

  \multirow{3}{*}{$s_{\bm{y}}$} & 1 &\multirow{3}{*}{7} & \multirow{3}{*}{7} & \multirow{3}{*}{38}  &  0.93 & 0   & 0.0070       & \multirow{3}{*}{23}   & \multirow{3}{*}{20000}& 34\\
 &  5&  &    &   &    0.95 & 0   & 0.0070  & &   & 6\\ 
 &  10&   &   &   &     1.00 & 0   & 0.0000  & &  & 6\\ 

\hline

\multirow{3}{*}{$s_{\bm{x}}$} & 1 &\multirow{3}{*}{80} & \multirow{3}{*}{60}   &  \multirow{3}{*}{241}  &  0.99 &  0 & 0.0004 &    \multirow{3}{*}{10} & \multirow{3}{*}{20000} & 33333 \\
 & {5} &   &    &   &   1.00 &  0 & 0.0002 &     & & 10 \\

 & 10 &   &    &    & 1.00 &  0 & 0.0000 &     &  & 10 \\

\hline

\multirow{3}{*}{$s_{\bm{d}}$} &  1 &\multirow{3}{*}{80} & \multirow{3}{*}{60}   &  \multirow{3}{*}{241}& 0.99 &  0 &  0.0004    &     \multirow{3}{*}{10}  & \multirow{3}{*}{20000} &  36363\\
&  5 &  &  &    & 1.00 &  0 &  0.0000    &       &  &  10\\ 
&  10 &  &    &  & 1.00 &  0 &  0.0000    &      &  &  10\\

\hline
\multirow{3}{*}{$s_{\bm{y}}$} &   1 &\multirow{3}{*}{80} & \multirow{3}{*}{60}   &  \multirow{3}{*}{241}&  0.98 & 0   & 0.0004        & \multirow{3}{*}{22}    & \multirow{3}{*}{20000} & 21\\
& 5 &  & &  &     1.00 & 0   & 0.0000        &     & & 7\\

& 10 &  & &     &  1.00 & 0   & 0.0000        &     & & 7\\

\hline
\multirow{3}{*}{$s_{\bm{x}}$} & 1 & \multirow{3}{*}{200} & \multirow{3}{*}{150}   &  \multirow{3}{*}{601}   & 1.00 & 0  & 0.0002 &    \multirow{3}{*}{10} & \multirow{3}{*}{25000} & $3.75 \times 10^5$\\ 
  & {5} &   &   &     & 1.00 & 0  & 0.0000 &    &  & 8\\ 

  & {10} &   &   &     & 1.00 & 0  & 0.0000 &    &  & 8\\ 

\hline
\multirow{3}{*}{$s_{\bm{d}}$} & 1&  \multirow{3}{*}{200} & \multirow{3}{*}{150}   &  \multirow{3}{*}{601} & 0.99 & 0   &  0.0002    &     \multirow{3}{*}{10}  & \multirow{3}{*}{25000} &  $4.06 \times 10^5$\\ 
 & 5 &  &  &      & 1.00 & 0   &  0.0000    &       &  &  8\\ 

 & 10 &  &  &      & 1.00 & 0   &  0.0000    &    & &  8\\ 

\hline

\multirow{3}{*}{$s_{\bm{y}}$} & 1 &  \multirow{3}{*}{200} & \multirow{3}{*}{150}   &  \multirow{3}{*}{601} &  0.99 &  0  & 0.0002        & \multirow{3}{*}{10}    & \multirow{3}{*}{25000}  & 37\\
& 5&  &  &      &  1.00 &  0  & 0.0000        &    & & 7\\
& 10&  &  &      &  1.00 &  0  & 0.0000        &     & & 7\\

\bottomrule
\end{tabular}
\caption{Numerical results for the facility location problem with $k \geq 1$ using OPT.}
\label{table:facility_multiple}
\end{table*}

\begin{table*}\centering
\footnotesize
\begin{tabular}{ccccccccccc}
\toprule
  Target & $k$ &  $n$     &$\Gamma$ & Accuracy   & Infeasibility & $sub_{max}$ &  $\abs{\mathcal{S}}$  &  $N$ & $t_{ratio}$    \\ 
\midrule

\multirow{3}{*}{$s_{\bm{x}}$} & 1 &\multirow{3}{*}{25} & \multirow{3}{*}{10}   & 1.00 & 0   & 0.0004  & \multirow{3}{*}{12} & \multirow{3}{*}{40000} & 1220 \\ 

  & 5    &    &    & 1.00 &  0 & 0.0000 &     &  & 7\\ 
  & 10  &   &     & 1.00 &  0 & 0.0000 &     & & 7\\

\hline

\multirow{3}{*}{$s_{\bm{d}}$} &  1 &\multirow{3}{*}{25} & \multirow{3}{*}{10} & 1.00 & 0   & 0.0004    & \multirow{3}{*}{28}   & \multirow{3}{*}{40000} & 1126 \\ 
&  5 &    &  & 1.00 &  0 &  0.0000    &       &  &  7 \\ 
&  10 &      &  & 1.00 &  0 &  0.0000    &    &  &7  \\ 

\hline

\multirow{3}{*}{$s_{\bm{y}}$}  & 1  &  \multirow{3}{*}{25} & \multirow{3}{*}{10} &  0.99 & 0   & 0.0004       & \multirow{3}{*}{30} & \multirow{3}{*}{40000} & 38\\
 &  5&  &   &     0.99 & 0   & 0.0002  &   & & 7\\
 & 10 &  &   &     0.99 & 0   & 0.0002 & & & 7\\

\bottomrule
\end{tabular}
\caption{Numerical results for the inventory control problem with $k \geq 1$ using OPT.}
\label{table:inventory_multiple}
\end{table*}

\begin{table*}\centering
\footnotesize
\begin{tabular}{cccccccccccc}
\toprule 
  Target & $k$ & $n$   & $m$  &$\Gamma$ &  Accuracy & Infeasibility & $sub_{max}$ &  $\abs{\mathcal{S}}$  &  $N$ & $t_{ratio}$    \\ 
\midrule

 \multirow{3}{*}{$s_{\bm{x}}$} &  1& \multirow{3}{*}{10} & \multirow{3}{*}{24}   & \multirow{3}{*}{0.1}  &  0.97 & 0   & 0.0010  & \multirow{3}{*}{17} &  20000 & 88137\\ 
& 5 &  & &   &  0.98 &  0 & 0.0004 &   & 20000 & 27\\ 

& 10 &  &  &  &  1.00 & 0  & 0.0002 &   & 20000 & 27\\

\hline

\multirow{3}{*}{$s_{\bm{d}}$} &   1&\multirow{3}{*}{10} & \multirow{3}{*}{24}   & \multirow{3}{*}{0.1} & 0.97 & 0   & 0.0010      & \multirow{3}{*}{24}   &  20000& $1.30 \times 10^5$ \\ 
&   5&  &  &    & 0.98 &  0 &  0.0004    &      & 20000 &  25\\ 
&  10&  &  &   &    0.99 & 0   &  0.0002    &      & 25000 &  25\\ 

\hline

\multirow{3}{*}{$s_{\bm{y}}$} & 1&  \multirow{3}{*}{10} & \multirow{3}{*}{24}   & \multirow{3}{*}{0.1}&  0.96 & 0   & 0.0010       & \multirow{3}{*}{32}   & 20000& 119\\
 &  5&  & &  &     0.98 & 0   & 0.0004       &     & 20000& 5\\

 & 10 &  &  &  &     0.99 &  0  & 0.0002        &     & 25000& 5\\

\bottomrule
\end{tabular}
\caption{Numerical results for the unit commitment problem with $k \geq 1$ using OPT.}
\label{table:UC_multiple}
\end{table*}

\paragraph{Results}
\begin{itemize}
    \item Both OPT and XGBoost consistently demonstrate excellent accuracy, never falling below 0.93 and often reaching 0.99 or 1.00. This performance holds true regardless of the problem size or the size of the uncertainty sets. Even when the solutions are not exactly accurate, the maximum suboptimalities remain exceptionally low, at most 0.001. This indicates the high quality of the solutions.
    \item The predictions are never infeasible for both OPT and XGBoost.
    \item In general, the prediction accuracy for the here-and-now decisions is the highest, followed by the worst-case scenarios and the wait-and-see decisions.
    \item  The solve times using OPT and XGBoost are significantly faster than Algorithm \ref{alg1}, at times reaching up to 3.37 million times faster. Additionally, OPT tends to outperform XGBoost in terms of speed. This is primarily because the time required for a decision tree to compute its predictions is typically less than a millisecond, whereas XGBoost generally takes slightly longer.  
    \item The speed-up of our approach to compute the wait-and-see decisions is less drastic compared to here-and-now decisions or worst-case scenarios, typically ranging from tens to hundreds of times faster. To compute a here-and-now decision or a worst-case scenario, the only computation needed is to determine the output of the trained model on an input. In order to compute a wait-and-see decision, however, we still need to solve a linear optimization problem, leading to longer computation time.
    \item OPT and XGBoost show very similar performance in general. This implies that we often do not have to compromise performance too much to gain interpretability.
    \item As we increase $k$, the quality of the solutions improves monotonically. At the same time, the relative speed-up of our approach decreases due to the evaluation process required to choose the best strategy.

\end{itemize}

\subsection{Analysis of Algorithm \ref{alg_partition}}
\label{sec:numerical_partition}

\begin{table*}% [H] is so declass\'e!
\centering
\footnotesize
\begin{minipage}{0.45\textwidth}
\begin{tabular}{cccccc}
\toprule
  $n$ & $m$ & $\Gamma$  & $\abs{\bm{\tau}}$ & $K$   & $|\bar{u}| - |\tau_{\bm{y}}|$       \\ 
\midrule
 7& 7 & 38& 22 &  1 &  9  \\
\hline
 80& 60 & 241 & 65&  13 &  9  \\
\hline 
200& 150& 751& 17498& 1& 148\\
\hline 
200& 150 & 601& 115& 1&12 \\
\bottomrule
\end{tabular}
\caption{Numerical results of Algorithm \ref{alg_partition} applied to the facility location problem.}
\label{table:facility alg_partition}
\end{minipage}\hfill
\begin{minipage}{0.45\textwidth}
\begin{tabular}{ccccc}
\toprule
  $n$ & $\Gamma$ & $\abs{\bm{\tau}}$ & $K$   & $|\bar{u}| - |\tau_{\bm{y}}|$      \\ 
\midrule
25 & 10 & 47&  7 &  9 \\
\hline
600 & 10& 90&  13 & 13  \\
\hline 
1000 & 10 &48& 5& 8\\
\hline 
1000 & 45&5550& 1&24 \\
\bottomrule
\end{tabular}
\caption{Numerical results of Algorithm \ref{alg_partition} applied to the inventory control problem.}
\label{table:inventory alg_partition}
\end{minipage}\par
\vskip\floatsep% normal separation between figures
\begin{tabular}{ccccccc}
\toprule
 $n$ & $m$ & $\Gamma$ & $\abs{\bm{\tau}}$ & $K$   & $|\bar{u}| - |\tau_{\bm{y}}|$       \\ 
\midrule
 10&24 & 0.1& 1492&  30 & 430 \\
\hline
10&24 & 2& 10482&  1 & 446  \\
\bottomrule
\end{tabular}
\caption{Numerical results of Algorithm \ref{alg_partition} applied to the unit commitment problem.}
\label{table:uc alg_partition}
\end{table*}

% \begin{table*}\centering
% \small
% \ra{1.3}
% \begin{tabular}{cccccc}
% \toprule
%   $n$ & $m$ & $\Gamma$  & $\abs{\bm{\tau}}$ & $K$   & $|\bar{u}| - |\tau_{\bm{y}}|$       \\ 
% \midrule
%  7& 7 & 38& 22 &  1 &  9  \\
% \hline
%  80& 60 & 241 & 65&  13 &  9  \\
% \hline 
% 200& 150& 751& 17498& 1& 148\\
% \hline 
% 200& 150 & 601& 115& 1&12 \\
% \bottomrule
% \end{tabular}
% \caption{Numerical results of Algorithm \ref{alg_partition} applied to the facility location problem.}
% \label{table:facility alg_partition}
% \end{table*}

% \begin{table*}\centering
% \small
% \ra{1.3}
% \begin{tabular}{ccccc}
% \toprule
%   $n$ & $\Gamma$ & $\abs{\bm{\tau}}$ & $K$   & $|\bar{u}| - |\tau_{\bm{y}}|$      \\ 
% \midrule
% 25 & 10 & 47&  7 &  9 \\
% \hline
% 600 & 10& 90&  13 & 13  \\
% \hline 
% 1000 & 10 &48& 5& 8\\
% \hline 
% 1000 & 45&5550& 1&24 \\
% \bottomrule
% \end{tabular}
% \caption{Numerical results of Algorithm \ref{alg_partition} applied to the inventory control problem.}
% \label{table:inventory alg_partition}
% \end{table*}

% \begin{table*}\centering
% \small
% \ra{1.3}
% \begin{tabular}{ccccccc}
% \toprule
%  $n$ & $m$ & $\Gamma$ & $\abs{\bm{\tau}}$ & $K$   & $|\bar{u}| - |\tau_{\bm{y}}|$       \\ 
% \midrule
%  10&24 & 0.1& 1492&  30 & 430 \\
% \hline
% 10&24 & 2& 10482&  1 & 446  \\
% \bottomrule
% \end{tabular}
% \caption{Numerical results of Algorithm \ref{alg_partition} applied to the unit commitment problem.}
% \label{table:uc alg_partition}
% \end{table*}

In this section, we demonstrate the effectiveness of Algorithm \ref{alg_partition}. We apply Algorithm \ref{alg_partition} in the previously described experiment, in case the number of distinct strategies for the wait-and-see decisions is extremely large. Tables \ref{table:facility alg_partition},  \ref{table:inventory alg_partition}, \ref{table:uc alg_partition} contain the numerical results on the facility location, the inventory control and the unit commitment problem, respectively.

\paragraph{Table Notations}

We use $\abs{\bm{\tau}}$ to denote the number of distinct tight constraints found in the training set, before applying Algorithm \ref{alg_partition}. As before, $K$ denotes the number it is reduced to. We also report how many more constraints the union constraints contain compared to individual tight constraints, denoted as $|\bar{u}| - |\tau_{\bm{y}}|$. Other columns are given to specify which problem Algorithm \ref{alg_partition} is applied to.

\paragraph{Results}
\begin{itemize}
        \item  When the number of distinct tight constraints found in the training instances is excessively large, we can combine the entire tight constraints into a single union constraints. In other words, the value of $K$ is set to 1. See Table \ref{table:facility alg_partition}, for example. In the facility location problem with $n= 200, m = 150, \Gamma = 751$, the entire set of 17498 tight constraints are combined to a single union constraints. Nevertheless, the increase in the number of constraints is relatively small, regarding that the total number of constraints in the deterministic version of this problem is 30350.  This result applies similarly to other examples with $K = 1$ as well. Therefore, Algorithm \ref{alg_partition} may not add too much additional computational burden even in extreme cases. 
        \item We have shown in Section \ref{sec:numerical_acc} that the prediction accuracy for the wait-and-see decisions is very high, even after we apply Algorithm \ref{alg_partition} to the training instances. This result holds true regardless of the value of $K$. This implies that even after reassigning the prediction targets of the training data, accurate machine learning models can still be trained.
\end{itemize}

{\color{black}
\subsection{Solving ARO with Varying Dimensions}
\label{sec:numerical_variability}

We evaluate the performance of the generalized approach discussed in Section \ref{subsec:variability} for problems with varying dimensions. We conduct two experiments on the inventory control problem with $n = 25$ and $\Gamma = 10$. 

In the first experiment, we randomly generate five distinct contingencies, each removing a certain portion of the here-and-now decision variables. These contingencies simulate situations where specific ordering options are no longer available.

In the second experiment, we randomly generate five different contingencies, each removing certain non-negativity constraints on the wait-and-see variables. These contingencies simulate situations where certain orders can be canceled without incurring additional costs.

Other experimental setups are identical to the descriptions in Section \ref{sec:description} and \ref{sec:numerical_acc}. For both experiments, we use a single categorical feature to encode the type of contingency. This results in five distinct categorical values, each corresponding to a specific type of contingency. Moreover, the number of strategies for the wait-and-see variables in these experiments is substantially higher compared to the previous experiment on the same inventory control problem: 90 and 102 for the two experiments, respectively. Therefore, we use Algorithm \ref{alg_partition} with $K = 1$ in this section. The main results are presented in Tables \ref{table:contingency_variable} and \ref{table:contingency_const}.

\paragraph{Results}
\begin{itemize}
    \item The number of unique strategies identified in the training set is larger compared to the previous experiment (refer to Table \ref{table:inventory} for comparison). While the number of strategies for the wait-and-see variables might seem smaller, this is due to the use of Algorithm \ref{alg_partition}, as mentioned earlier. This increased diversity results from the existence of multiple contingencies.
    \item Our approach continues to achieve near-perfect performance, demonstrating its effectiveness for problems with varying contingencies.
\end{itemize}

\begin{table*}\centering
\footnotesize
\begin{tabular}{cccccc}
\toprule 
  Target & $k$   &  Accuracy & Infeasibility & $sub_{max}$ &  $\abs{\mathcal{S}}$      \\ 
\midrule

 \multirow{3}{*}{$s_{\bm{x}}$} &  1   &  0.99 & 0   & 0.0005  & \multirow{3}{*}{19} \\ 
& 5 &     1.00 &  0 & 0.0000 &       \\ 

& 10 &      1.00 & 0  & 0.0000 &      \\

\hline

\multirow{3}{*}{$s_{\bm{d}}$} &  1   &  0.98 & 0   & 0.0005  & \multirow{3}{*}{65} \\ 
& 5 &     0.99 &  0 & 0.0005 &       \\ 

& 10 &      1.00 & 0  & 0.0003 &      \\ 

\hline

\multirow{3}{*}{$s_{\bm{y}}$}  &  1   &  0.99 & 0   & 0.0005  & \multirow{3}{*}{19} \\ 
& 5 &     1.00 &  0 & 0.0000 &       \\ 

& 10 &      1.00 & 0  & 0.0000 &      \\

\bottomrule
\end{tabular}
\caption{Numerical results for the inventory control problem with varying number of decision variables.}
\label{table:contingency_variable}
\end{table*}

\begin{table*}\centering
\footnotesize
\begin{tabular}{cccccc}
\toprule 
  Target & $k$   &  Accuracy & Infeasibility & $sub_{max}$ &  $\abs{\mathcal{S}}$      \\ 
\midrule

 \multirow{3}{*}{$s_{\bm{x}}$} &  1   &  0.99 & 0   & 0.0004  & \multirow{3}{*}{19} \\ 
& 5 &     1.00 &  0 & 0.0000 &       \\ 

& 10 &      1.00 & 0  & 0.0000 &      \\

\hline

\multirow{3}{*}{$s_{\bm{d}}$} &  1   &  0.99 & 0   & 0.0004  & \multirow{3}{*}{61} \\ 
& 5 &     0.99 &  0 & 0.0004 &       \\ 

& 10 &      1.00 & 0  & 0.0004 &      \\ 

\hline

\multirow{3}{*}{$s_{\bm{y}}$} &  1   &  0.99 & 0   & 0.0004  & \multirow{3}{*}{19} \\ 
& 5 &     1.00 &  0 & 0.0000 &       \\ 

& 10 &      1.00 & 0  & 0.0000 &      \\

\bottomrule
\end{tabular}
\caption{Numerical results for the inventory control problem with varying number of constraints.}
\label{table:contingency_const}
\end{table*}

}

\subsection{Solving ARO with Suboptimal Strategies}
\label{sec:largescale}

In this section, we apply Algorithm \ref{alg:prescriptive} to solve large scale unit commitment problems using suboptimal strategies. The size of the unit commitment problem we consider is $n = 100$ and $m = 24$, which is much larger than the scale we considered in Section \ref{sec:numerical_acc}. Throughout the experiment, we set the optimality gap of Gurobi to 0.005. When generating a training set in Phase 1 by solving ARO instances, we set $\epsilon_1 = 0.05$ and $\epsilon_2 = 0.01$ for Algorithm \ref{alg1} and Algorithm \ref{alg2}, respectively. When computing suboptimalities to generate reward matrices and choose the best among $k>1$ predictions, we set $\epsilon_2 = 0.001$. When evaluating the final output of the decision trees to assess the ultimate effectiveness of our approach on the test set, we set $\epsilon_1 = 0.001, \epsilon_2 = 0.001$, for precise assessment. Moreover, the number of unique strategies in the training set is prohibitively large in this experiment, as we will demonstrate below. Therefore, we set $Q_1 = Q_2 = Q_3 = 40$. We perform a grid search over the maximum depths 5 and 10 for OPT. Table \ref{table:UC_largescale} contains the main experiment results. The notations are identical to the previous sections.

\paragraph{Results}

\begin{itemize}
    \item The accuracies are much lower compared to the previous results. However, the maximum suboptimalities are still around 0.02, indicating that the predictions are of reasonable quality. As in Section \ref{sec:numerical_acc}, the predictions are always feasible.
    \item When $k = 1$, the solve time using OPT can be more than 10 million times faster than Algorithm \ref{alg1} . This scale of speed-up is much more drastic than the previous results. However, as $k$ increases, the relative speed-up becomes similar to the previous results.
    \item The accuracies and maximum suboptimalities are identical across different prediction targets, and the performances do not improve as we increase $k$. 
    \item Overall, even if we use only 40 out of 3341 strategies found, Algorithm \ref{alg:prescriptive} can find high-quality solutions.
\end{itemize}

\begin{table*}\centering
\footnotesize
\begin{tabular}{cccccccccc}
\toprule 
  Target & $k$   &$\Gamma$ &  Accuracy & Infeasibility & $sub_{max}$ &  $\abs{\mathcal{S}}$  &  $N$ & $t_{ratio}$    \\ 
\midrule

 \multirow{3}{*}{$s_{\bm{x}}$} &  1 & \multirow{3}{*}{2}  &  0.18 & 0   & 0.0205  & \multirow{3}{*}{3341} &  \multirow{3}{*}{10000} & $8.49 \times 10^7$\\ 
& 5 &  &   0.18 &  0 & 0.0205 &     &  & 10 \\ 

& 10 &  &    0.18 & 0  & 0.0205 &    &  & 10\\

\hline

\multirow{3}{*}{$s_{\bm{d}}$} &   1 & \multirow{3}{*}{2}  &  0.18 & 0   & 0.0205  & \multirow{3}{*}{3341} &  \multirow{3}{*}{10000} & $7.07 \times 10^7$\\ 
& 5 &  &   0.18 &  0 & 0.0205 &     &  & 10\\ 

& 10 &  &    0.18 & 0  & 0.0205 &    &  & 10\\ 

\hline

\multirow{3}{*}{$s_{\bm{y}}$} & 1 & \multirow{3}{*}{2}  &  0.18 & 0   & 0.0205  & \multirow{3}{*}{3341} &  \multirow{3}{*}{10000} & 4871 \\ 
& 5 &  &   0.18 &  0 & 0.0205 &     &  & 10\\ 

& 10 &  &    0.18 & 0  & 0.0205 &    &  & 10\\

\bottomrule
\end{tabular}
\caption{Numerical results for the unit commitment problem with $n= 100, m = 24, k \geq 1$ using OPT.}
\label{table:UC_largescale}
\end{table*}

{\color{black}

\subsection{Analysis of Algorithm \ref{alg1_warm} and \ref{alg1_fixed}}
\label{sec:accel}

In this section, we assess the effectiveness of Algorithms \ref{alg1_warm} and \ref{alg1_fixed} using the unit commitment problem with $n = 100$ and $m = 24$.  We implement $\mathcal{I}_1$, $\mathcal{I}_2$, and $\mathcal{I}_3$ as feedforward neural networks with two hidden layers, each consisting of 32 neurons. These models are trained using the Adam optimizer \citep{KingBa15} with a learning rate of 0.001, implemented in PyTorch \citep{DBLP:journals/corr/abs-1912-01703}. We set $\epsilon_1 = 0.05$ and $\epsilon_2 = 0.01$ for all algorithms. We compare the solution outputs and the runtime of Algorithm \ref{alg1} with those of Algorithm \ref{alg1_warm} and \ref{alg1_fixed} on 200 test instances. Consistent with our previous implementation, we use a random $\bm{x}_0$ for Algorithm \ref{alg1} and three random initial points for Algorithm \ref{alg2}. We vary the size of the training data for the intermediate models $\mathcal{I}_1$, $\mathcal{I}_2$, and $\mathcal{I}_3$ to observe any performance changes. We use relatively smaller training data compared to Algorithms \ref{alg:overview} and \ref{alg:prescriptive} to demonstrate the effectiveness of Algorithms \ref{alg1_warm} and \ref{alg1_fixed} even with a limited dataset. We report the experiment results in Table \ref{table:accel}. 

\paragraph{Table Notations}
In the columns Algorithm and $N'$, we report the type of acceleration algorithm used (Algorithms \ref{alg1_warm} or \ref{alg1_fixed}) and the number of training samples for $\mathcal{I}_1$, $\mathcal{I}_2$, and $\mathcal{I}_3$, respectively. In the Proportion column, we report the proportion of binary variables in the test set that are fixed or warm-started using the $p$ values decided in the validation set (recall that $p$ is the threshold value for the probability output of $\mathcal{I}_1$ to decide which entries of the output will be used). In the $t_{{ratio}}$ column, we report the relative speed-up compared to Algorithm \ref{alg1}, rounded to the second decimal place. In the $sub_{{max}}$ column, we report the maximum suboptimality of the solution output compared with the solution output of Algorithm \ref{alg1}. Note that unlike the experiments in Section \ref{sec:numerical_acc}, the relative speed-up and the suboptimality are computed with respect to Algorithm \ref{alg1} and \ref{alg2} with $\epsilon_1 = 0.05$ and $\epsilon_2 = 0.01$, not the near-optimal version with very tight tolerances.

\begin{table*}\centering
\footnotesize
\begin{tabular}{ccccc}
\toprule 
  Algorithm & $N^{'}$   & Proportion &  $sub_{max}$ & $t_{ratio}$  \\ 
\midrule

Algorithm \ref{alg1_warm} &  \multirow{2}{*}{1000} & \multirow{2}{*}{0.22}  &  0.0000 & 2.89   \\ 
Algorithm \ref{alg1_fixed}&  &  &   0.0001 &  7.56  \\

\hline
Algorithm \ref{alg1_warm} &  \multirow{2}{*}{2000} & \multirow{2}{*}{0.51}  &  0.0000 & 3.25   \\ 
Algorithm \ref{alg1_fixed}&  &  &   0.0001 &  8.95  \\

\hline
Algorithm \ref{alg1_warm} &  \multirow{2}{*}{3000} & \multirow{2}{*}{0.71}  &  0.0000 & 3.45   \\ 
Algorithm \ref{alg1_fixed}&  &  &   0.0000 &  9.19  \\

\hline
Algorithm \ref{alg1_warm} &  \multirow{2}{*}{4000} & \multirow{2}{*}{0.88}  &  0.0000 & 4.28   \\ 
Algorithm \ref{alg1_fixed}&  &  &   0.0000 &  11.72  \\

\bottomrule
\end{tabular}
\caption{Numerical results of Algorithm \ref{alg1_warm} and \ref{alg1_fixed} applied to the unit commitment problem.}
\label{table:accel}
\end{table*}

\paragraph{Results}
\begin{itemize}
    \item As expected, Algorithm \ref{alg1_fixed} outperforms both Algorithm \ref{alg1_warm} and \ref{alg1} in terms of speed. Specifically, with just 4000 training data, Algorithm \ref{alg1_fixed} achieves more than a 10-fold speedup compared to Algorithm \ref{alg1}, while Algorithm \ref{alg1_warm} achieves more than a 4-fold speedup.
    \item The maximum suboptimality of the solutions is practically negligible across all training data sizes. This indicates that the solutions generated by Algorithm \ref{alg1_warm} and \ref{alg1_fixed} are virtually identical to those generated by Algorithm \ref{alg1}.
    \item As the number of training data increases, the proportion of variables that can be fixed or warm-started also increases, indicating an improvement in the accuracy of $\mathcal{I}_1$. For $N^{'} = 4000$, approximately 88\% of the binary variables can already be fixed or warm-started on average. Naturally, as this proportion increases, the solution speed of Algorithm \ref{alg1_warm} and \ref{alg1_fixed} also improves.
\end{itemize}

}

\section{Conclusions}
\label{sec:conclusion}

Despite the theoretical advantages of ARO compared to RO, existing solution algorithms generally suffer from heavy computational burden. We proposed a machine learning approach to solve two-stage ARO with polyhedral uncertainty sets and binary here-and-now variables. We generate multiple ARO instances by varying a key parameter of the problem, and solve them with Algorithm \ref{alg1}. Using the parameters as features, we train a machine learning model to predict high-quality strategies for the here-and-now decisions, the worst-case scenarios associated with the here-and-now decisions, and the wait-and-see decisions. {\color{black}We also proposed learning-based algorithms to expedite training data generation, and} a partitioning algorithm to reduce the number of distinct target classes to make the prediction task easier. Numerical experiments on synthetic and real-world problems show that our approach can find high quality solutions of ARO problems significantly faster than the state-of-the-art algorithms.

\section*{Acknowledgements}
The research was partially supported by a grant to MIT from Lincoln Laboratories.

%% If you have bibdatabase file and want bibtex to generate the
%% bibitems, please use
%%
\bibliographystyle{model5-names} 
\bibliography{reference}

@article{Lefebvre2024,
	author = {Lefebvre, Henri and Malaguti, Enrico and Monaci, Michele},
	journal = {INFORMS Journal on Computing},
	number = {1},
	pages = {78-96},
	title = {Adjustable Robust Optimization with Discrete Uncertainty},
	volume = {36},
	year = {2024}}

@article{cauligi2021coco,
	author = {Cauligi, A. and Culbertson, P. and Schmerling, E. and Schwager, M. and Stellato, B. and Pavone, M.},
	journal = {IEEE Robotics and Automation Letters},
	number = {2},
	year = {2022},
	pages = {1447--1454},
	title = {CoCo: Online {Mixed}-{Integer} {Control} via {Supervised} {Learning}},
	howpublished = {https://doi.org/10.1109/LRA.2021.3135931},
	volume = {7},
}

@misc{nair2021solving,
      title={Solving Mixed Integer Programs Using Neural Networks}, 
      author={Vinod Nair and Sergey Bartunov and Felix Gimeno and Ingrid von Glehn and Pawel Lichocki and Ivan Lobov and Brendan O'Donoghue and Nicolas Sonnerat and Christian Tjandraatmadja and Pengming Wang and Ravichandra Addanki and Tharindi Hapuarachchi and Thomas Keck and James Keeling and Pushmeet Kohli and Ira Ktena and Yujia Li and Oriol Vinyals and Yori Zwols},
      year={2021},
      eprint={2012.13349},
      archivePrefix={arXiv},
      primaryClass={math.OC}
}

@article{doi:10.1287/ijoc.2016.0723,
	author = {Alvarez, Alejandro Marcos and Louveaux, Quentin and Wehenkel, Louis},
	journal = {INFORMS Journal on Computing},
	number = {1},
	pages = {185-195},
	title = {A Machine Learning-Based Approximation of Strong Branching},
	volume = {29},
	year = {2017}}

@article{Balcan_Sandholm_Vitercik_2020,
	author = {Balcan, Maria-Florina and Sandholm, Tuomas and Vitercik, Ellen},
	journal = {Proceedings of the AAAI Conference on Artificial Intelligence},
	month = {Apr.},
	number = {04},
	pages = {3227-3234},
	title = {Learning to Optimize Computational Resources: Frugal Training with Generalization Guarantees},
	volume = {34},
	year = {2020}}

@InProceedings{KingBa15,
  author    = {Kingma, Diederik and Ba, Jimmy},
  booktitle = {International Conference on Learning Representations (ICLR)},
  title     = {Adam: A Method for Stochastic Optimization},
  year      = {2015},
  address   = {San Diega, CA, USA},
  optmonth  = {12},
}

@article{DBLP:journals/corr/abs-1912-01703,
  author       = {Adam Paszke and
                  Sam Gross and
                  Francisco Massa and
                  Adam Lerer and
                  James Bradbury and
                  Gregory Chanan and
                  Trevor Killeen and
                  Zeming Lin and
                  Natalia Gimelshein and
                  Luca Antiga and
                  Alban Desmaison and
                  Andreas K{\"{o}}pf and
                  Edward Z. Yang and
                  Zach DeVito and
                  Martin Raison and
                  Alykhan Tejani and
                  Sasank Chilamkurthy and
                  Benoit Steiner and
                  Lu Fang and
                  Junjie Bai and
                  Soumith Chintala},
  title        = {PyTorch: An Imperative Style, High-Performance Deep Learning Library},
  journal      = {CoRR},
  volume       = {abs/1912.01703},
  year         = {2019},
  url          = {http://arxiv.org/abs/1912.01703},
  eprinttype    = {arXiv},
  eprint       = {1912.01703},
  bibsource    = {dblp computer science bibliography, https://dblp.org}
}

@article{Kai2023,
	author = {Wang, Kai and Aydemir, Mehmet and Jacquillat, Alexandre},
	journal = {INFORMS Journal on Optimization},
	number = {0},
	pages = {null},
	title = {Scenario-Based Robust Optimization for Two-Stage Decision Making Under Binary Uncertainty},
	volume = {0},
	year = {2023}}

@article{GOERIGK2023529,
	author = {Marc Goerigk and Mohammad Khosravi},
	journal = {European Journal of Operational Research},
	number = {2},
	pages = {529-551},
	title = {Optimal scenario reduction for one- and two-stage robust optimization with discrete uncertainty in the objective},
	volume = {310},
	year = {2023}}

@article{COHEN202383,
	author = {Izack Cohen and Krzysztof Postek and Shimrit Shtern},
	journal = {European Journal of Operational Research},
	number = {1},
	pages = {83-104},
	title = {An adaptive robust optimization model for parallel machine scheduling},
	volume = {306},
	year = {2023}}

@inproceedings{Sun2014Daily,
  title={Adaptive robust optimization for daily power system operation},
  author={Sun, A. and Lorca, A.},
  booktitle={2014 Power Systems Computation Conference (PSCC)},
  pages={1--9},
  year={2014},
  organization={IEEE}
}

@article{BertsimasSun,
  title={Adaptive robust optimization for the security constrained unit commitment problem},
  author={Bertsimas , D. and Litvinov , E. and Sun , A. and Zhao , J. and Zheng , T.},
  journal={IEEE Transactions on Power Systems},
  volume={28},
  number={1},
  pages={52--63},
  year={2013},
  publisher={IEEE}
}

@article{Policy,
author = {Amram, M. and Dunn, J. and Zhuo, Y. D.},
  title     = {Optimal Policy Trees},
  journal   = {Machine Learning},
  volume    = {11},
  year      = {2022},
  pages = {2741-2768}
 }

@article{Sun2014Wind,
author = {Sun, A. and Lorca, A.},
title = {Adaptive robust optimization with dynamic uncertainty sets for multi-period economic dispatch under significant wind},
year = {2015},
journal = {IEEE Transactions on Power Systems},
volume = {30},
issue = {4},
pages = {1702--1713},
publisher={IEEE}
}

@article{UCdata,
    author = {Carrion, M. and Arroyo, J.M.},
    title = {A computationally efficient mixed-integer linear formulation for the thermal unit commitment problem},
    journal = {IEEE Transactions on Power Systems},
    year = {2006},
    volume = {21},
    pages = {1371--1378}
}

@article{OPT,
author = {Bertsimas, Dimitris and Kim, Cheol Woo},
title = {A Prescriptive Machine Learning Approach to Mixed-Integer Convex Optimization},
journal = {INFORMS Journal on Computing},
year = {2023},
doi = {10.1287/ijoc.2022.0188},

URL = { 
    
        https://doi.org/10.1287/ijoc.2022.0188
    
    

},
eprint = { 
    
        https://doi.org/10.1287/ijoc.2022.0188
    
    

}
}

@article{Voice,
  title={The voice of optimization},
  author={Bertsimas, D. and Stellato, B.},
 journal={Machine Learning},
volume={110},
  year={2021},
  pages={249-277}
}

@article{Prune,
  title={Online mixed-integer optimization in milliseconds},
  author={Bertsimas, D. and Stellato, B.},
  journal={INFORMS Journal on Computing},
  volume ={34}, 
  number ={4}, 
  year={2022}
}

@article{survey,
author = {Yanıkoğlu, \.{I}. and Gorissen, B. and den Hertog, D.},
year = {2019},
month = {09},
pages = {799-813},
title = {A Survey of Adjustable Robust Optimization},
volume = {277},
journal = {European Journal of Operational Research},
doi = {10.1016/j.ejor.2018.08.031}
}

@article{inventory1,
author = {See, C. and Sim, M.},
year = {2009},
title = {Robust Approximation to Multiperiod Inventory Management},
volume = {58},
number = {3},
pages = {583-594},
journal = {Operational Research},
}

@article{inventory2,
author = {Ang, M. and Lim, Y. and Sim, M.},
year = {2012},
title = {Robust Storage Assignment in Unit-Load Warehouses},
volume = {58},
number = {11},
pages = {2114-2130},
journal = {Management Science},
}

@article{portfolio,
title = {Adjustable robustness for multi-attribute project portfolio selection},
journal = {European Journal of Operational Research},
volume = {252},
number = {3},
pages = {931-946},
year = {2016},
issn = {0377-2217},
doi = {https://doi.org/10.1016/j.ejor.2016.01.058},
url = {https://www.sciencedirect.com/science/article/pii/S0377221716300017},
author = {Thomas Fliedner and Juuso Liesiö}
}

@misc{gurobi,
  author = {{Gurobi Optimization, LLC}},
  title = {{Gurobi Optimizer Reference Manual}},
  year = 2023,
  url = "https://www.gurobi.com"
}

@article{adaptive,
  title={Adjustable robust optimization via fourier-motzkin elimination},
  author={Zhen, J. and den Hertog, D. and Sim, M.},
  journal={Operations Research},
  volume={66},
  number={4},
  pages={1086--1100},
  year={2018}
}

@inproceedings{XGBoost,
 author = {Chen, T. and Guestrin, C.},
 title = {{XGBoost}: A Scalable Tree Boosting System},
 booktitle = {Proceedings of the 22nd ACM SIGKDD International Conference on Knowledge Discovery and Data Mining},
 series = {KDD '16},
 year = {2016},
 isbn = {978-1-4503-4232-2},
 pages = {785--794},
 numpages = {10},
 publisher = {ACM},
}

@book{robusttext,
  title={Robust Optimization},
  author={Ben-Tal, A. and Nemirovski, A. and Laurent El, G.},
  year={2009},
  publisher={Princeton University Press}
}

@book{Bertsimasrobusttext,
  title={Robust and Adaptive Optimization},
  author={Bertsimas, D. and den Hertog, D.},
  year={2022},
  publisher={Dynamic Ideas}
}

@article{CCG,
  title={Solving two-stage robust optimization problems using a column-and-constraint generation method},
  author={Zeng, B. and Zhao, L.},
  journal={Operations Research Letters},
  volume={41},
  number={5},
  pages={457--461},
  year={2013}
}

@article{adaptivefirst,
  title={Adjustable robust solutions of uncertain linear programs},
  author={Ben-Tal, A. and Goryashko, A. and  Guslitzer, E. and  Nemirovski, A.},
  journal={Mathematical Programming},
  volume={99},
  pages={351 -- 376},
  year={2004},
}

@misc{InterpretableAI,
  author = "Interpretable AI, LLC",
  title = "Interpretable AI Documentation",
  year = 2023,
  url = "https://www.interpretable.ai"
}

\end{document}